\documentclass[a4paper,conference]{IEEEtran}

\ifCLASSINFOpdf
   \usepackage[pdftex]{graphicx}
\else
\fi
\usepackage{amsmath,amssymb}

\usepackage{comment}
\usepackage{color}

\usepackage{graphbox,graphicx, caption}
\usepackage{soul}
\usepackage{verbatim}
\usepackage{bm}
\usepackage{booktabs}
\usepackage{hyperref}
\usepackage{xspace}

\newcommand{\eg}{\emph{e.g.} }

\newcommand{\txt}{\mathbf{x_R}}
\newcommand{\img}{\mathbf{x_I}}

\newcommand{\txtFeat}{\mathbf{q_R}}
\newcommand{\imgFeat}{\mathbf{q_I}}

\newcommand{\ItoA}{\texttt{PITA}\xspace}

\newcommand{\Real}{{\mathbb{R}}}

\begin{document}
\title{Picture-to-Amount (PITA): Predicting Relative Ingredient Amounts from Food Images 
%\hl{use acronym PITA: Picture-to-amount?}
}

\author{\IEEEauthorblockN{Jiatong Li\IEEEauthorrefmark{1},
Fangda Han\IEEEauthorrefmark{1},
Ricardo Guerrero\IEEEauthorrefmark{2} and
Vladimir Pavlovic\IEEEauthorrefmark{1}\IEEEauthorrefmark{2}}
\IEEEauthorblockA{\IEEEauthorrefmark{1}Department of Computer Science, Rutgers University,
Piscataway, NJ, USA}
\IEEEauthorblockA{\IEEEauthorrefmark{2}Samsung AI Center, Cambridge, UK}}

\maketitle

\begin{abstract}
Increased awareness of the impact of food consumption on health and lifestyle today has given rise to novel data-driven food analysis systems. Although these systems may recognize the ingredients, a detailed analysis of their amounts in the meal, which is paramount for estimating the correct nutrition, is usually ignored. 
% In this paper, we study the novel and challenging problem of predicting the relative amount of each ingredient needed to prepare the observed food item from an image of that item. 
In this paper, we study the novel and challenging problem of predicting the relative amount of each ingredient from a food image. 
We propose \ItoA, the Picture-to-Amount deep learning architecture to solve the problem. More specifically, we predict the ingredient amounts using a domain-driven Wasserstein loss from image-to-recipe cross-modal embeddings learned to align the two views of food data. Experiments on a dataset of recipes collected from the Internet show the model generates promising results and improves the baselines on this 
%exceptionally difficult
challenging task. A demo of our system and our data is available at: foodai.cs.rutgers.edu.

\end{abstract}


% \fi
\IEEEpeerreviewmaketitle

\section{Introduction}
Increased awareness of the impact of food consumption on health and lifestyle today has given rise to novel data-driven food analysis systems, e.g.,~\cite{beijbom2015menu,meyers2015im2calories}, whose goals are to 
% alleviate the challenge of 
find effective and efficient ways for 
tracking daily food intake and, subsequently, enable automatic dietary assessment or even facilitate possible positive changes in lifestyle. Many of those systems use data modalities such as images to seamlessly extract information related to the food item that was consumed, 
% often
where the information includes
the identity of the meal, its ingredients, or even its caloric value. While these systems frequently claim to predict the energy intake, they base these predictions on standard energy tables of standardized ingredients (e.g., USDA\footnote{https://ndb.nal.usda.gov/ndb/search/list}). The estimation of the food amount, a highly challenging and often ambiguous task, is delegated to the users themselves. Even systems that aim to predict a fine-grained ingredient-based representation of the food item, e.g., \cite{salvador2017learning}, \cite{chen2016deep}, do not consider the problem of predicting the ingredient amounts or relative contributions of ingredients in each dish. However, these amounts are paramount for estimating the correct energy value of the meal. A small amount of high-fat food might not be a major health risk factor, while an unhealthy ingredient with a dominant amount may lead to potential health problems. Analyzing individual nutrients of a food dish in detail and how each ingredient contributes to health through its 
%energy calorie proxy
nutritions can resolve the immediate and important goal of dietary assessment. Estimating the fine-grained ingredient
%-calorie
information can also facilitate choices of which ingredients could be replaced %with their low-calorie counterparts or to substitutions 
according to users' dietary needs, leading to overall improvements in nutritional behavior. Therefore, we argue there is a need for food analysis systems which output fine-grained ingredient amounts.

In this paper, we study the novel problem of predicting the relative amount of each ingredient in a food item from images, where relative amounts are the proportions of each ingredient needed to prepare the food, relative to the total ingredient weight. %They are called "relative" because s
Scaling all ingredient amounts to the same factor will not change the food or recipe itself. Only the number of servings will be changed. 

\begin{figure*}[htbp]
    \centering
    \includegraphics[width=0.7\linewidth]{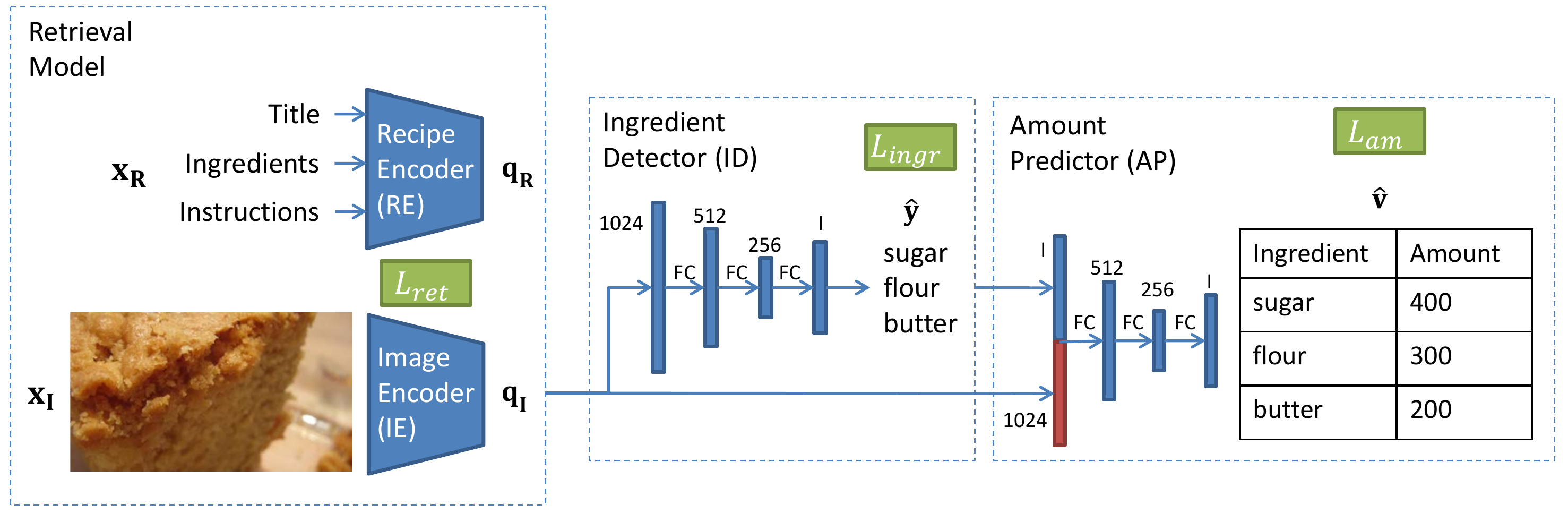}
    \caption{Overview of our \ItoA approach. We first learn an embedding representation from retrieval system. We proceed to detect the ingredients from embeddings and finally predict amounts from embeddings and detected ingredients. %\hl{Like this:  Here is an example of a system diagram, which I find created well. } \url{http://geometry.cs.ucl.ac.uk/projects/2020/relate} 
    %\hl{It is simple, balanced, uses mathematical notation, and the notation is the same as the one used in the paper (i.e., functions and variables).}
    }
    \label{fig:overview}
\end{figure*}

% There are two main challenges associated with ingredient amount predictions. 
Several critical challenges are associated with ingredient amount predictions. \textbf{First}, some ingredients, like salt, are invisible. The existence and amounts of these ingredients can only be inferred through correlation with other ingredients. However, these ingredients cannot be simply ignored in food analysis systems. They are important in the cooking process as they impart specific flavors to the dish, while at the same time they can be crucial in estimating nutritional values, %or health impact,
c.f.,~\cite{he2009comprehensive}. %In Recipe1M dataset \cite{salvador2017learning}, salt is one of the most frequent ingredients, included in about half of the recipes. 
% Meanwhile, too much sodium in diet can lead to high blood pressure \cite{he2009comprehensive}. 
%Salt plays an important role as a flavor enhancer during cooking, nonetheless, too much sodium in diet can lead to high blood pressure \cite{he2009comprehensive}, while at the same time might ruin a dish's taste.
Therefore, it is essential to consider invisible ingredients.
%in food analysis systems. 
\textbf{Second}, cooking involves complex operations that transform ingredients both mechanically and chemically. Attributes such as shape and color can change drastically during this process. 
%Ingredients might be diced, sliced or pureed; powdered ingredients (\eg  cake mix) might become solid; meat will change color during cooking, with its final color depending on the cooking method and the mixing with other ingredients.Analyzing the amounts of ingredients may also suffer from occlusion.
\textbf{Last}, the ingredients are very diverse and distributed extremely unevenly in different dishes, leading to high-dimensional and highly unbalanced estimation problems.
% To the computer vision community, the problem involves both detection and regression at the same time. 
% Furthermore, the items to detect demonstrate large intra-class variation and they might not have specific shapes.

% To summarize, invisibility, large intra-class variation as well as highly unbalanced distribution all make the ingredients amounts estimation a tough problem to solve.

To tackle these challenges, we propose the \ItoA framework. More specifically, we aim at first learning an embedding representation of the image from a cross-modal image-to-text retrieval system, such as~\cite{han2019art}. When learning the retrieval model, ingredient co-occurrences and cooking actions are encoded by the text encoder. This information is related to the image appearance through the image encoder and the shared image-text features. As illustrated in \autoref{fig:overview}, embeddings are used to condition 
ID and, in a cascade, AP. In this manner, the model drives the cross-modal embedding to be highly relevant to the task of estimation, while being recursively conditioned from simpler (detection) to more complex tasks (estimation). 
%Therefore during detection, the invisible ingredients can be inferred through correlation with other ingredients. 
The unbalanced problem can be addressed by the positive sample weight in the ID loss. During estimation, the relationship of amounts are learned by AP using Wasserstein distance, the transport loss which ideally matches our problem because ingredients can be classified into functional groups like spices and oil.
% different similarities between ingredients should be considered. T
The penalty of predicting vegetable oil as olive oil should be less than that of predicting meat as vegetables. The ingredient functional substitution groups are first selected by Word2vec-like~\cite{mikolov2013distributed} cosine similarities which are trained on titles, ingredients and instructions of Recipe1M \cite{salvador2017learning}. Ingredient pairs above a certain threshold are proposed, then curated by human annotators. The distance matrix of the Wassterstein transport is determined by the substitution groups and the cosine distances of Word2vec embeddings between ingredients. When constructing the functional groups, the dimensions are also reduced. Canonical ingredient construction reduces the number of total ingredients (output dimension) from 16k in Recipe1M to 1.4k. The ingredients form 172 substitution groups, which further reduces the complexity and enhances generalization of the approach. Experiments on a dataset of recipes collected from the Internet show the model generates state-of-the-art results, improving previous baselines.

In summary, the contributions of this paper are:
\begin{itemize}
    \item We propose a novel and challenging problem: analyzing the relative amounts of the ingredients from a food image.
    \item We define a novel semi-automatic process to create ingredient substitution groups and facilitate data-driven functional ingredient equivalencies.
    \item We propose \ItoA deep learning framework to solve the problem of relative ingredient-specific amount estimation and improve previous baselines.
\end{itemize}

\section{Related Work}
% As food plays an essential part in our life, t
A growing body of research in computer vision has focused on the problems of computational food classification, cross-modal retrieval, and ingredient analysis. While prior work has  addressed those difficult problems, more demanding tasks of food volume and food amount estimation, needed for detailed nutritional analysis, are emerging as the key topics in computational food analysis and challenges for computer vision and machine learning communities.

\textbf{Food Classification}. Hand-crafted features and traditional classifiers were initially used to classify food images into food categories or meal types, such as vegetables, soups, fish, grilled fish, etc.  Both \cite{beijbom2015menu} and  \cite{bossard2014food} used SVMs as classifier. \cite{bossard2014food} leveraged super-pixels in combination with random forests, while \cite{beijbom2015menu} employed a bag-of-visual-words approach. \cite{beijbom2015menu} also estimated the energy caloric content of a meal from a restaurant menu based on the classification results.
Classical computer vision features tend to be outperformed by modern methods that rely on deep learning for feature extraction or directly using deep learning end-to-end.  \cite{ciocca2017food} and \cite{wang2015recipe} used deep learning features while \cite{kagaya2014food}, \cite{singla2016food}, \cite{merler2016snap}, \cite{chen2017chinesefoodnet} and \cite{mezgec2017nutrinet} trained deep-learning models for food recognition. \cite{martinel2018wide} introduced slice convolution blocks and \cite{qiu2019mining} adversarially mines discriminative food regions for food classification.

%\subsection{Cross-modal Retrieval}
%  In general terms, c
 \textbf{Cross-modal Retrieval}. %Instead of solving the classification problem, some approaches focus on 
 %the task of 
 %retrieving relevant data across modalities.
 In the context of the work proposed here, this means that given an image, the system retrieves its recipe from a collection of test recipes or vice-versa. 
%  \hl{VP: I would not talk about data first, describe the approach then maybe mention the data.  Why is the data even important at this point?} Recipe1M\cite{salvador2017learning} is a dataset built with retrieval in mind, with Recipe1M+\cite{marin2019recipe1m+} expanding it
% %  the dataset Recipe1M\cite{salvador2017learning} 
%  from 800k images to 13M. However, this expansion is based on image web search using recipe titles as queries. Therefore, it is likely that a large proportion this expanded set of images 
%  no longer offer true cross-modal representations of their assigned recipes.
% %  no longer share the same ingredients or instructions as the original ones. 
%  Recipe1M+ also expands by providing annotations on the amount and nutritional information for a subset of 50k recipes, while \cite{li2019deep} annotate 250k recipes from Recipe1M, with both nutritional and ingredient amount information, out of which 80k also contain images. 
%  Using Recipe1M, 
 \cite{chen2017cross} proposed a framework that learns a joint embedding of recipe ingredients and images while \cite{salvador2017learning} leveraged titles, ingredients and instructions of recipes. In AdaMine \cite{carvalho2018cross}, recipes from the same semantic class were embedded closer. \cite{chen2018deep} incorporated attention mechanism to the recipe embedding %a cross-modal retrieval model 
 in order to focus on words that facilitate retrieval. ACME \cite{wang2019learning} imposed modality alignment and cross-modal translation consistency using an adversarial learning strategy to improve the retrieval performance. 
 Although cross-modal retrieval can be applied to the ingredient amount estimation problem by returning the ingredients and amounts of the best matching recipe in the database of recipes, this approach cannot readily generalize to novel foods and highly depends on the collection of recipes.

%\subsection{Ingredient Analysis}
 \textbf{Ingredient Analysis}. A significantly more challenging task is predicting the existence of ingredients given a food image. \cite{chen2016deep} used multi-task deep learning and a graph modeling ingredient co-occurrences. \cite{salvador2019inverse} predicted ingredients as sets and generated cooking instructions by attending to both image and its inferred ingredients simultaneously. \cite{min2019ingredient} developed an ingredient-guided cascaded multi-attention network for ingredient detection. 
However, these works do not tackle the quantity prediction problem, which is significantly more difficult than that of identifying the ingredients alone.

%\subsection{Volume Estimation}
 \textbf{Volume Estimation}. Estimating the total food amount can be addressed in the context of multi-view 3D reconstruction or single-view volume estimation.
Multi-view reconstruction approaches~\cite{shashua1994relative,dehais2017two} largely rely on on traditional computer vision techniques.% to accomplish this task.
~\cite{zheng2018multi} estimated the contour of the food from three different views and matched it with a predefined library to estimate the volume. However, their library used only nine types of food with over-simplified shapes. Our work does not depend on any predefined library and only needs images and ingredient amounts.
%\hl{VP: need to say briefly what it does, not just how good it is}, which performs well in very constrained data instances  \hl{VP: "but the performance might come from an easy dataset" need to say why the dataset is easy}. 
\cite{liang2017computer} used faster-RCNN to detect food and applied graph cuts to find the contour to estimate volume. However, food images in their dataset are well-balanced and easily segmentable. Compared with them, real-world datasets have a highly unbalanced distribution of ingredients, dominated by meal images with a complex mix of many ingredients.
% deep learning in their work, it's \hl{VP: don't use abbreviated forms when writing: it's should be written it is} only for object detection and they use a simple method to calculate volume. \hl{VP: again, you do not really mention the key information of what method they use nor why you judge it as simple.  Need to justify your claims!}

In the significantly more challenging single view setting, \cite{meyers2015im2calories} aimed to recognize the contents of a meal first then predict the calories for home-cooked foods. Their method is based on deep learning for classification and depth estimation to calculate the volume and calories. \cite{fang2018single} used GANs, however their approach requires densely annotated datasets, which are not available for real-world data. The systems in \cite{ege2019image} either uses size-known reference objects, including rice grains, or special devices, like inertial sensors and stereo cameras.

Estimating relative ingredient amounts, on the other hand, does not require reference objects or special devices. Methods estimating relative ingredient amounts can be applied to food images acquired in uncalibrated settings.%, where camera information and size-known reference objects are not available.
~Furthermore, while previous work calculated the amounts of
% foods 
the meal
as a whole, estimating relative amounts of individual ingredients provides important additional information that can be linked to knowledge sources such as the micronutrition, wellness impact of individual food categories, etc. \cite{li2019deep} was the first to attempt to predict relative ingredient amounts with CNNs. However, their methods were not significantly better than the retrieval baselines. In this work, we leverage retrieval-inspired cross-modal image-text embeddings and minimize the Wasserstein distance of the food amount transport, driven by domain-specific ingredient subgroups. Experiments show that this approach can lead to new state-of-the-art on this challenging task.

\section{Methods}
\label{sec:method}
Ingredient relative amount prediction is a novel and ambiguous task. We first introduce some notations and define the problem in \autoref{sec:definition}. We start with the im2recipe retrieval system in \autoref{sec:retrieval}. As the problem involves both ingredient detection and estimation, we detect the ingredients from embeddings as in \autoref{sec:id} and predict the amounts given the embeddings and the detected ingredients in \autoref{sec:amount}. \autoref{fig:overview} illustrates our approach. 

\subsection{Notation and Problem Definition}
\label{sec:definition}

%\hl{I do not think the variable names should use more than one symbol, with possible subscripts or superscripts.  That's a standard mathematical notation.  You can use multiple symbols for abbreviations of elements of system.  So, I would use $x_I$ and $x_R$ for image and recipe inputs. Then, $v$ for amount (volume), $y$ for ingredient identity vector instead of $ID_y$, $z_I$ and $z_R$ for embeddings of images and recipes, etc.  There is no need to define $v_x$ and $v_y$, there is only one $v$ and the predicted amount will be $\hat{v}$, unless you want to say that you are predicting amounts from images and recipes, in which case it would be $v_I$ and $v_R$ for GT amounts and the hat notation for predicted.  Or you can use the subscript 'GT' to denote the GT variable, e.g., $v_{GT}$ would be the ground truth relative amount.  The 'hat' notation is preferred. }
\begin{itemize}
    \item {\bf Relative Amount.} 
    %\hl{I would use the term ``normalized'', as it more closely reflects what we actually use; but we can stick with ``relative'' as well.} 
    Given recipe $\mathbf{x_R}$, we define the {\bf relative amount vector} 
    $\mathbf{v}=(v_{1},\ldots,v_{i},\ldots,v_{I}) \in \Real_{\geq0}^{I}$, where $I$ is the total number of ingredients and $v_{i} = \frac{C}{M} m_{i}$ is the relative amount of the $i-th$ ingredient, with $m_i \geq 0$ the absolute mass of ingredient $i$ in grams, $M = \sum_{i}^I m_i$ the total mass of all ingredients, and $C$ a normalizing constant. For convenience, we set $C = 1000gr$. $v_{i}=0$ when the $i-th$ ingredient is not present in the recipe. 
    %The values in each amount vector sum up to $C$ grams.  \hl{Is there even any need to define the absolute amount (vector)?  It is never used, we only use relative amounts.  So I would just immediately and only define the relative amount.}
    % Suppose $a$ is a constant and $a>0,~a\mathbf{v_y}$ also represents the same recipe. Therefore we assume that $\mathbf{v_y}$ is normalized such that $\sum_{i=1}^Iv_{y_i}=C$ where $C$ is a constant. 
    %So $v_{y_i}$ can be interpreted as the proportion of the $i-th$ ingredient in the recipe, or there are $v_{y_i}$ grams of the $i-th$ ingredient in one unit of recipe (\eg $C$ grams).
    
    \item {\bf Ingredient Detection.} Given recipe $\mathbf{x_R}$, define the {\bf Ingredient Detection Vector} $\mathbf {y}=\mathbf{1}_{\{\mathbf{v}>0\}} \in \{ 0, 1\}^I$.
    
    \item {\bf Ingredient Distance.} Let $\bm{M}\in [0,1]^{I\times I}$ be a matrix derived from substitution groups. $M_{ij}=M_{ji},~M_{ij}\geq 0, M_{ii}=0.$ For ingredient triplet $(i,j,k)$, if ingredient $j$ is more similar to ingredient $i$ than ingredient $k$ determined by substitution groups, $M_{ij}<M_{ik}.$ The matrix $\bm{M}$ is called {\bf Ingredient Distance Matrix}. 
    
    \item {\bf Predicted quantities.} We use the ``hat'' notation to refer to predicted quantities.  For instance, $\mathbf{\hat{v}}=(\hat{v}_{1},\ldots,\hat{v}_{i},\ldots,\hat{v}_{I})$ refers to the amount vector predicted from an image. %\in [0,C]^I$ such that $\sum_{i=1}^Iv_{x_i}=C.$
    %Correspondingly, the ingredient detection vector of the image is $\mathbf{\hat{y}}=\mathbf{1}_{\{\mathbf{\hat{v}}>0\}}$.

\end{itemize}

\subsection{Retrieval Model}
\label{sec:retrieval}
% \begin{figure}
%     \centering
%     \includegraphics[width=1.0\linewidth]{cvpr2020AuthorKit/latex/figures/retrieval.png}
%     \caption{Retrieval model.} %\hl{VP: is this figure even necessary?  Why not just say you use the system from im2recipe paper and put a reference?JL: it's not the same as im2recipe or Fangda's WACV one} }%\hl{VP: this figure has very different style from fig.1.  We should make sure all figs use the same style.} }
%     \label{fig:retrieval}
% \end{figure}

The retrieval model is based on \cite{han2019art} that embeds a pair of text and image into a shared Food Space. The model is trained to maximize the similarity between a \textbf{positive} pair (\eg text and image come from the same recipe), meanwhile minimizing that between a \textbf{negative} pair (\eg text and image come from different recipes).

Formally, the retrieval model takes a pair of text (including title, ingredients and instructions) and image $(\txt, \img)$ as input. \textbf{Text} $\txt$ is first trained with a Word2vec model that embeds each word into a vector. \textbf{Title} is treated as one sentence and is forwarded through an LSTM. The final hidden status is used as the feature of title. \textbf{Ingredients} go through a bidirectional LSTM to mitigate the influence of initial order as ingredients should be unordered. \textbf{Instructions} are more complicated because they usually contain several sentences that frequently exceed a hundred words. To effectively remember the long-term dependency in instructions, a two-layer LSTM structure is applied; the first layer is used to encode each step of the instructions into a vector. These "step vectors" are forwarded through another LSTM  as a sequence and the final hidden state is used as the representation of the instructions. The features from title, ingredients and instructions are concatenated and passed through a fully-connected layer to yield the output $\mathbf{z_R}$ of the Text Encoder, 
\begin{equation}
    \mathbf{z_R} = F_{txt}\left( \txt \right).    
\end{equation}
\textbf{Image} is encoded by a ResNet50~\cite{he2016deep}. The result after average pooling is used as the the initial feature and is passed through a fully-connected layer to get the output of IE. The output image feature $\mathbf{z_I}$ has the same dimension as the text feature above. This process is encapsulated in the following model:
\begin{equation}
    \mathbf{z_I} = F_{img}\left( \img \right).  
\end{equation}
Based on the assumption that text features and image features share the Food Space, a fully-connected layer is applied to both text and image features. The mapping uses a weight sharing mechanism to further align the features from the two modalities in the same space: %\hl{Is FC really the same exact network, with identical parameters, for both image and text modalities?}
\begin{align}
    \txtFeat &= F_{fc} \left( \mathbf{z_R} \right), \\
    \imgFeat &= F_{fc} \left( \mathbf{z_I} \right),
\end{align}
where $\txtFeat$ and $\imgFeat$ are the final text feature and image feature in the Food Space.

\textbf{Loss} is designed to minimize the similarity between a positive pair and maximize that between a negative pair. Cosine similarity is applied. 
%A naive way to compute the loss is to find all similarities in one batch and average them. However, during training, the model is easily learned from trivial pairs, which makes a large fraction of pairs useless in one batch. Therefore, focusing on those `painful pairs' could potentially lead to better performance.
We applied hard-sample mining~\cite{hermans2017defense} to make the model focus on the most difficult pairs. The loss $L_{ret}$ in one batch is defined as
\begin{align}
    L_{ret} &= \frac{1}{N}\sum_{i=1}^N \max \left(0, m - \cos(\txtFeat_i, \imgFeat_i) + \cos(\txtFeat_i, \mathbf{u}_j)  \right), \nonumber\\
    i,j &\in [1,N] \; and \; i\ne j,
\end{align}
% \hl{VP: is there a hinge function here?}
where $N$ is the batch size, $m$ is the margin between positive pair similarity and negative pair similarity ($m$ is set to $0.3$ by cross validation), $\cos(\txtFeat_i, \imgFeat_i)$ is the cosine similarity between a positive pair and $\cos(\txtFeat_i, \mathbf{u}_j)$ is that between a negative pair. Hard-sample mining is applied here to choose the most difficult sample $\mathbf{u}_j$ (could be text or image), which is defined as the most similar negative sample with $\txtFeat_i$.

\subsection{Ingredient Detection}
\label{sec:id}
As the problem ingredient relative amount prediction involves both ingredient detection and estimation, after getting the image embedding $\imgFeat$ from the retrieval system in \autoref{sec:retrieval}, we first do ingredient detection using %a network parametrized by $\theta_{ID}$ to guide the amount regression. 
an ingredient detector, as in \autoref{fig:overview}.
The soft probability prediction is 
\begin{equation}
    %\mathbf{p_x}=f_{\theta_{ID}}(\mathbf{q_x}),
    \mathbf{p}=F_{id}(\mathbf{q_I}).
\end{equation}
%where $f_{\mathbf{\theta_{ID}}}(\cdot)$ is ID in \autoref{fig:overview} and 
$p_{i}\in [0,1], i=1,2,\ldots,I$. %$I$ is the total number of ingredients. \hl{Already defined that previously} %\hl{So if you are using the notation that the subscript of a function name denotes the function parameter, as you did here, you should then be consistent and use it for all functions.  Else, put the parameter as the conditioning variable:  $f_{ID}(q|\theta_{ID})$.  But, again, be consistent and use the same notation everywhere.}

We minimize the positive sample weighted binary cross entropy loss between $\mathbf{p_x}$ and $\mathbf{y}$. More specifically, 
\begin{equation}
    L_{ingr}=-\sum_{i=1}^I[w_iy_i\log p_{i}+(1-y_i)\log (1-p_{i})],
\end{equation}
where $w_i$ is the positive sample weight, which is essential as ingredients are sparse.% Take Recipe1M \cite{salvador2017learning} as an example, the average ratio of negative samples to positive samples over all ingredients, calculated by
%\begin{equation}
%    \bar{r}=\frac{1}{I}\sum_{i=1}^I\frac{\sum_{\mathbf{y} \in R}(1-ID_{y_i})}{\sum_{\mathbf{y} \in R}ID_{y_i}},
%\end{equation}
% where $R$ is the collection of recipes, is over 2000, which means on average there is less than one appearance of ingredient $i$ in 2000 recipes.

To stabilize training, we set
\begin{equation}
   w_i=\min(t,\frac{\sum_{r \in R}(1-y_{r_i})}{\sum_{r \in R}y_{r_i}}),
\end{equation}
 where $R$ is the collection of recipes in the training set, $r$ is a recipe and $t$ is a threshold. In our experiment, $t$ is set to 4 by cross validation.% to ensure the average number of predicted ingredients is around the average number of ground truth ingredients.
 
 The hard prediction is the soft probability prediction thresholded to 0.5,
 \begin{equation}
     \mathbf{\hat{y}}=\mathbf{1}_{\{\mathbf{p}>0.5\}}.
 \end{equation}
$\mathbf{\hat{y}}$ is then passed to AP in \autoref{fig:overview} to guide the amount prediction.

\subsection{Amount Prediction}
\label{sec:amount}
Given the image embedding $\mathbf{q_x}$ from \autoref{sec:retrieval} and ingredient detection vector $\mathbf{\hat{y}}$ from \autoref{sec:id}, we train %a network with parameters $\mathbf{\theta_{AM}}$ to predict the amounts. 
an amount predictor, as in \autoref{fig:overview}.
More specifically, 
\begin{equation}
    %\mathbf{v_x}=f_{\mathbf{\theta_{AM}}}(\mathbf{q_x}\oplus\mathbf{ID_x})\odot \mathbf{ID_x},
    \mathbf{\hat{v}}=F_{ap}(\mathbf{q_I}\oplus\mathbf{\hat{y}})\odot \mathbf{\hat{y}},
\end{equation}
where %$f_{\mathbf{\theta_{AM}}}(\cdot)$ is AP in \autoref{fig:overview}, 
$\oplus$ means concatenation and $\odot$ denotes element-wise multiplication.  

 We minimize the Wasserstein distance \cite{frogner2015learning} between $\mathbf{\hat{v}}$ and $\mathbf{v}.$ The loss function is 
\begin{equation}
    L_{am}=W_p^p(\mathbf{\hat{v}},\mathbf{v},\bm{M}),%=\inf_{\mathbf{T}\in \Pi(\mathbf{v_x},\mathbf{v_y})}<\mathbf{T},\mathbf{M}>,
    \label{eq:loss}
\end{equation}
% \hl{VP: always need comma before where, which, etc. The comma needs to be in the equation that precedes this. Fix in whole document.} 
%where $\mathbf{T}$ denotes the transportation between $\mathbf{v_x}$ and $\mathbf{v_y}$ and $\Pi(\mathbf{v_x},\mathbf{v_y})=\{\mathbf{T} \in R_+^{I\times I}: \mathbf{T1}=\mathbf{v_x}, \mathbf{T}^T\mathbf{1}=\mathbf{v_y}\}$. $\mathbf{1}$ is the all-one vector. Equation \eqref{eq:loss} 
% \hl{VP: use \textbackslash eqref\{eq:loss\} command. Fix everywhere.} 
where %$\mathbf{v_x},\mathbf{v_y}$ are predicted and ground truth amounts respectively \hl{but you previously defined $v_y$ as the absolute amount in III.A} and 
$\bm{M}$ is the ingredient distance matrix defined in \autoref{sec:definition}. %When calculating the loss, $\mathbf{\hat{v}}$ and $\mathbf{v}$ are linearly scaled such that their L1 norm, or the total weights of the ingredients, are the same.
Equation \eqref{eq:loss} can be minimized with the method in \cite{frogner2015learning}.
 
\section{Experiments}
\subsection{Implementation Details}
The architectures of ID and AP are shown in \autoref{fig:overview}. Both are fully connected networks
%. %The ingredient detection network takes the 1024-dimensional image embedding as input while the amount prediction network takes the concatenation of image embedding and the predicted amounts ($1024+I$-dimensional) as input. 
% For both networks, the layers are activated 
with leaky ReLU except the output layers. The negative slope of the leaky ReLU is 0.2. The output layer of ID is activated with sigmoid while the output layer of AP is activated with softmax.
% \begin{figure}
%     \centering
%     \begin{minipage}[t]{.45\linewidth}
%     \includegraphics[width=1.0\linewidth]{cvpr2020AuthorKit/latex/figures/id_network.png}
%     \subcaption{Ingredient Detection Network.}
%     \label{fig:id network}
%     \end{minipage}
%     \begin{minipage}[t]{.45\linewidth}
%     \includegraphics[width=1.0\linewidth]{cvpr2020AuthorKit/latex/figures/amount_network.png}
%     \subcaption{Amount Prediction Network.}
%     \label{fig:amount network}
%     \end{minipage}
%     \caption{Ingredient Detection and Amount Prediction Network Architectures. \hl{VP: can you move this into Fig. 1? Fig.1 is a bit coarse on its own.}}
% \end{figure}

The retrieval system, ID and AP are all optimized with Adam optimizer. The learning rate is $10^{-4}$ and the batch size is 64. The three parts are learned sequentially in three incremental stages: 1) retrieval system, 2) ingredient detection, and 3) amount prediction.  We find that fine tuning previous parts using weighted sum loss does not significantly improve performance. When training the retrieval system, we follow \cite{salvador2017learning} and \cite{han2019art} where the ingredient amounts are not used. During test time, the image is first passed through IE to get an embedding representation $\mathbf{q_x}$. Then $\mathbf{q_x}$ is passed through ID to get $\mathbf{\hat{y}}$. Finally $\mathbf{q_x}$ and $\mathbf{\hat{y}}$ are passed through AP to get the amounts.
%and introduces more tuning hyperparameters.

\subsection{Dataset and Metrics}
Our dataset is based on Recipe1M \cite{salvador2017learning} which includes over one million recipes scraped from the Internet including titles, ingredients descriptions with units and amounts (1 cup of flour) and instructions. About 400k of the recipes have images. For amount analysis, we use the 80k subset of Recipe1M annotated for amounts and with images in \cite{li2019deep}. 

% \noindent
\textbf{Ingredients}. As the number of unique ingredients identified in Recipe1M is over 16k including single and plural forms (egg, eggs) and nearly the same ingredient with different names (msg, accent\_seasoning), we use the canonical ingredient construction in \cite{li2019deep}, which reduces the number of ingredients to 1.4k with 95\% coverage for a more compact ingredient list. After removing the non-food ingredients like tin\_foils and skewers, the number of ingredients is 1362. ID and AP are trained with the 48k training data of the annotated recipes while the retrieval system is trained with the training part of recipes with amounts and the rest of recipes with images that are not annotated for amounts, about 371k in total.

% \noindent
\textbf{Ingredient Substitution Groups}. The 1.4k ingredient list still contains ingredient substitutes like (butter, margarine) which should not be penalized as much as predicting butter as apples. Therefore, we construct ingredient substitution groups both for learning and evaluation. We first select ingredient pairs whose cosine similarity in Word2vec embedding space, which is trained on the instructions, ingredients and titles of Recipe1M, is greater than 0.6, resulting in five neighbors per ingredient on average. Human annotators then approve or reject the selected pairs. They can also add more pairs according to ingredient ontology, like (strawberries, mixed\_fruits). 3142 out of 6342 ingredient pairs are accepted by human annotators. Next, we treat ingredients as nodes and approved or added pairs as edges of a graph. We compute the connected components of the graph and each connected component is one substitution group. This results in 172 substitution groups. An example of an ingredient group created is: duck fat, chicken fat, pork fat, bacon drippings and lard; that is, forms of animal fat.  The ingredient distance matrix $\bm{M}$ is then calculated as follows:
\begin{itemize}
    \item Ingredient $i$, $j$ belong to the same substitution group: $M_{ij}=0$.
    \item $i,~j$ belong to different substitution groups: $M_{ij}=1-cos(\mathbf{em_i},\mathbf{em_j})$, where $cos(\cdot,\cdot)$ is cosine distance and $\mathbf{em_i}$ is the Word2vec embedding of ingredient $i$. %In other words, if two ingredients belong to different substitution groups, their bin distance is the cosine distance of their Word2vec embeddings.
\end{itemize}
We report three evaluation metrics, the first two evaluate ingredient detection and the last evaluates amounts, all in the context of substitution groups. The number of common ingredients $c$ between $\mathbf{y},~\mathbf{\hat{y}}$ is

\noindent %$c=\min(\mathbf{ID_x}^T\mathbf{1}_{\{\bm{S}\cdot\mathbf{ID_y}>0\}},\mathbf{ID_y}^T\mathbf{1}_{\{\bm{S}\cdot\mathbf{ID_x}>0\}})$.
 $c=\sum_{g_i}\min(\sum_{k\in g_i}y_k,\sum_{k\in g_i}\hat{y}_k)$ where $g_i$ is a substitution group and $k$ is an ingredient.

The metrics are:
\begin{itemize}
    \item {\bf CVG:} The coverage of ground truth ingredient, calculated by $\frac{c}{\sum_{i=1}^Iy_i}$.
    \item {\bf IOU:} Intersection over the union of ground truth and detected, calculated by $\frac{c}{\sum_{i=1}^Iy_i+\sum_{i=1}^I\hat{y}_i-c}$
    \item {\bf EMD:} Earth mover's distance between ground truth and predicted amounts, using $\bm{M}$ as the distance matrix. %Both amount vectors are linearly scaled such that the ingredient amounts sum to $C=1000$ grams.
\end{itemize}

We also report group level metrics. The distance between two substitution groups is the cosine distance between group centroids in Word2vec embedding space. Suppose the group level distance matrix is $\bm{M_1}$. For substitution groups $g_i,~g_j,~M_{1_{ij}}=1-cos(\frac{1}{|g_i|}\sum_{k\in g_i}\mathbf{em_k},\frac{1}{|g_j|}\sum_{k\in g_j}\mathbf{em_k})$ where $|g_i|,|g_j|$ are the sizes of the substitution groups and k is an ingredient. The ingredient detection and amount vectors of the ground truth are collapsed according to groups. More specifically, suppose there are $g$ substitution groups. Define the group-ingredient matrix as $\bm{G}\in \{0,1\}^{g\times I}, G_{ik}=1$ if ingredient $k$ belongs to group $g_i$. The collapsed amount vectors are $\mathbf{v_{1}}=\bm{G}\cdot\mathbf{v},~\mathbf{\hat{v}_{1}}=\bm{G}\cdot\mathbf{\hat{v}}$ and the collapsed ingredient detection vectors are $\mathbf{y_{1}}=\mathbf{1}_{\{\bm{G}\cdot\mathbf{y}>0\}},~\mathbf{\hat{y}_{1}}=\mathbf{1}_{\{\bm{G}\cdot\mathbf{\hat{y}}>0\}}.$ The group level metrics are:
\begin{itemize}
    \item {\bf CVG (Group):} Group-level coverage, calculated by $\frac{\mathbf{y_{1}}^T\mathbf{\hat{y}_{1}}}{\sum_{i=1}^Iy_{1_i}}$.
    \item {\bf IOU (Group):} Group-level intersection over union, calculated by
    
    $\frac{\mathbf{y_{1}}^T\mathbf{\hat{y}_{1}}}{\sum_{i=1}^Iy_{1_i}+\sum_{i=1}^I\hat{y}_{1_i}-\mathbf{y_{1}}^T\mathbf{\hat{y}_{1}}}$.
    \item {\bf EMD (Group):} Group-level earth mover's distance between $\mathbf{v_{1}}$ and $\mathbf{\hat{v}_{1}}$, using $\bm{M_1}$ as the distance matrix. %Both amount vectors are linearly scaled such that the ingredient amounts sum to $C=1000$ grams.
\end{itemize}
Specifically, {\bf CVG (Group)} and {\bf IOU (Group)} can be viewed as calculated the usual way where substitutions are not considered and there are only 172 ingredients.

\subsection{Compared Methods}
We investigate both retrieval-based and prediction-based methods, as cross-modal retrieval can be seen as a baseline method to ingredient amount prediction by returning the ingredients and amounts of the retrieved recipe. Both ingredient-level and group-level metrics are reported.

As for prediction-based methods, we study both ingredient-level and group-level methods. Ingredient-level methods predict detailed amounts of all 1362 ingredients while group-level methods treat each substitution group as one ingredient.
%and only output amounts of the whole substitution group. 
For ingredient-level methods, both ingredient-level and group-level metrics are reported. For group-level methods, only group-level metrics are reported.

The compared retrieval-based methods are:
\begin{itemize}
    %\item {\bf Retrieval-w/ gt:}  This is the performance upper-bound as it has access to the ground truth during test phase.
    \item {\bf Retrieval:} Use only the retrieval model in \autoref{sec:retrieval}. Given an image, retrieve the top-1 recipe from a collection of 1000 randomly chosen recipes in the test set, following \cite{salvador2017learning}. The ground truth recipe is not in the collection.\footnote{If the ground truth recipe is included in the collection, it will be the performance upper-bound as the model has access to the ground truth during test phase. The upper bound is: CVG$=0.70$, IOU$=0.59$, EMD$=119.20$, CVG(Group)=$0.71$, IOU(Group)=$0.61$, EMD(Group)=$237.90$.}
    \item {\bf ATTEN\cite{chen2018deep}:} The method in \cite{chen2018deep}.
    \item {\bf ACME\cite{wang2019learning}:} The method in \cite{wang2019learning}.
\end{itemize}
We compare the retrieval model with \cite{chen2018deep} and \cite{wang2019learning} because their methods are the most similar to ours in terms of model architecture and loss function. A detailed comparison of the retrieval models is in \autoref{tab:retrieval_methods}.
\begin{table*}[]
    \centering
    \caption{Retrieval-based Methods}
    %\begin{tabular}{p{1.8cm}p{1.5cm}p{2cm}p{2cm}p{1.5cm}p{1.5cm}}
    \begin{tabular}{ccccc}
    \toprule
    & Title Encoder & Ingredient Encoder & Instruction Encoder & Loss Function \\
    \midrule
    Retrieval & LSTM& bidirectional LSTM & two-layer LSTM  & hard mining triplet \\
    %Retrieval w/o gt & No & LSTM& bidirectional LSTM & two-layer LSTM  & hard mining triplet \\
    ATTEN\cite{chen2018deep} & LSTM& bidirectional LSTM & two-layer LSTM  & triplet \\
    ACME\cite{wang2019learning}  & - & bidirectional LSTM & skip-thought + LSTM & hard mining triplet \\
    %AdaMine\cite{10.1145/3209978.3210036} & - & bidirectional LSTM & skip-thought + LSTM & triplet+semantic \\
    %Im2recipe\cite{salvador2017learning} & No & - & bidirectional LSTM & skip-thought + LSTM & cosine + classification \\
    \bottomrule
    \end{tabular}
    \label{tab:retrieval_methods}
\end{table*}

For prediction-based methods, the compared ingredient-level methods are:
\begin{itemize}
    \item {\bf IE+AP:} The dense method in \cite{li2019deep}. The model uses a Resnet50 \cite{he2016deep} to predict the ingredient amounts. The output layer is activated by softmax.  The cross entropy between the ground truth and the predicted amounts is minimized. The result is thresholded to top-10 ingredients as the average number of ingredients in Recipe1M is about ten \cite{salvador2017learning}.
    \item {\bf IE+RE+AP(Wass):} The model only uses one network to predict the amounts from Food Space embeddings. Wasserstein distance between ground truth and predicted amounts using $\bm{M}$ as the distance matrix is minimized. The result is thresholded to top-10 ingredients.
    \item {\bf IE+RE+AP(CE):} The model architecture is the same as {IE+RE+AP(Wass)} but the loss function is cross-entropy as \cite{li2019deep}. The result is thresholded to top-10 ingredients. 
    \item {\bf IE+ID+AP:} The model separates ingredient detection and amount prediction. IE is pretrained on UPMC \cite{wang2015recipe} as in \cite{li2019deep} and its weights are learned when training ID. AP is trained as in \autoref{sec:amount}.
    \item {\bf IE+RE+ID+AP(Wass):} Our method in \autoref{sec:method}. 
    %\item {\bf IE+RE+ID+AP(L1):} The model architecture is the same as \autoref{fig:overview} but AP does not use substitution group information or $\bm{M}$ during training. The L1 loss $\sum_{i=1}^I|v_{y_i}-v_{x_i}|$ between $\mathbf{v_x},\mathbf{v_y}$ is minimized.
    \item {\bf IE+RE+ID+AP(CE):} The model architecture is the same as \autoref{fig:overview} but AP does not use substitution group information or $\bm{M}$ during training. The cross-entropy loss $\sum_{i=1}^Iv_i\log (\hat{v}_i+\epsilon)$ between $\mathbf{v},\mathbf{\hat{v}}$ is minimized. $\epsilon=10^{-6}$ is added for numerical stability as $\mathbf{\hat{v}}$ is sparse.
\end{itemize}

The group-level compared methods are:
\begin{itemize}
    \item {\bf IE+RE+AP(Wass)(group):} The network architecture is the same as {IE+RE+AP(Wass)} except that the output dimension of {IE+RE+AP(Wass)} is the number of ingredients 1362 while the output dimension of {IE+RE+AP(Wass)(group)} is the number of groups 172. Wasserstein distance between the collapsed ground truth and predicted amounts using $\bm{M_1}$ as distance matrix is minimized. The result is thresholded to top-10 groups as the average number of groups in ground truth is also about ten. This is probably because ingredients within a substitution group do not frequently co-occur.
    \item {\bf IE+RE+ID+AP(Wass)(group):} The group-level counterpart of {IE+RE+ID+AP(Wass)}. ID is trained with substitution group-level binary cross-entropy loss. The positive sample weight is $\min(t,\frac{\sum_{r \in R}(1-y_{1_{r_i}})}{\sum_{r \in R}y_{1_{r_i}}})$ and $t$ is set to 4. For AP, Wasserstein distance between ground truth and predicted amounts using $\bm{M_1}$ as distance matrix is minimized.
\end{itemize}
\subsection{Results and Evaluation}
The performance of the ingredient-level methods and the group-level methods are shown in Table \ref{tab:performance}. The numbers are "mean over test set". The up arrows indicate the higher the better and the down arrows indicate the lower the better. 
%We report both ingredient-level and group-level metrics for ingredient-level methods and only group-level metrics for group-level methods. 
The best method 
%except {\bf Retrieval w/gt} with respect to each metric 
is marked ${\mathbf {bold}}$.
%, as {\bf Retrieval w/gt} is the performance upperbound.
%\hl{VP: may be better to use colors, like red and blue} . 
\newcommand{\tableSecond}[1]{$\mathbf{#1}$}
\newcommand{\tableBest}[1]{$\mathbf{\underline{#1}}$}

\begin{table*}[]
    % \footnotesize
    \centering
    \caption{Performance of Compared Methods. %\hl{I wonder if we should be showing all these methods; what's the point of showing the ones that do not perform well?  Unless there is a specific reason for showing those.  The table looks busy and the text in related sections is overfilled with different model acronyms/abbreviations. I would, as a reader, find that confusing. For instance, IE+RE+ID+AP L1 and Wass perform almost equally well.  Do we need to show both?  %IE+RE+AP(CE) does quite well, but we do not have IE+RE+ID+AP(CE) evaluated; why?
    %} 
    }
    
     %\begin{tabular}{lccp{1.4cm}p{1.4cm}p{1.4cm}p{1.4cm}}
    %\begin{tabular}{lcccccc}
    \begin{tabular}{ccccccc}
    \toprule
    \textbf{Method} & \textbf{CVG$\uparrow$}&\textbf{IOU$\uparrow$}&\textbf{EMD$\downarrow$} & \textbf{CVG (Group)}&\textbf{IOU (Group)}&\textbf{EMD (Group)}\\
    \midrule
     Retrieval& $0.51$ & $0.34$& $191.13$ & $0.54$ & $0.38$ & $384.07$ \\
      %Retrieval w/ gt$^\text{best case bound}$ & \textcolor{red}{$0.70$}&  \textcolor{red}{$0.59$} & \textcolor{red}{$119.20$} & \textcolor{red}{$0.71$} & \textcolor{red}{$0.61$} &\textcolor{red}{$ 237.90$}\\
      ATTEN\cite{chen2018deep} &$0.47$&$0.32$&$205.19$&$0.50$&$0.35$&$407.73$\\
      %\cite{chen2018deep} w/ gt & $0.59$ & $0.49$ &$157.11$ &$0.62$&$0.51$&$311.09$\\
      ACME\cite{wang2019learning} &$0.48$&$0.33$&$199.87$&$0.51$&$0.36$&$399.90$\\
      %\cite{wang2019learning} w/ gt & & &&&&\\
      IE+AP & $ 0.45$ & $ 0.25$& $193.33$ & $0.46$ & $0.30$ & $ 396.52$\\
      IE+RE+AP(Wass) & $ 0.26$ & $ 0.13$ & \tableSecond{142.18}& $0.2$ & $0.22$ & $370.20$\\
      IE+RE+AP(CE) &$0.50$&$0.28$&$145.30$&$0.51$&$0.35$&$307.78$\\
      IE+ID+AP &$0.46$&$0.30$&$220.18$&$0.48$&$0.34$&$446.51$\\
      %IE+RE+ID+AP(L1)& \tableSecond{0.63} & \tableSecond{0.42} & $154.50$ & $0.65$ & \tableSecond{0.48} & $314.98$ \\
      IE+RE+ID+AP(CE)& \tableSecond{0.63} & \tableSecond{0.42} & $154.35$ & $0.65$ & \tableSecond{0.48} & $315.98$ \\
      IE+RE+ID+AP(Wass) & \tableSecond{0.63} & \tableSecond{0.42} & $147.29$ & $0.65$ & \tableSecond{0.48} & $319.63$\\
      \midrule
     IE+RE+AP(Wass)(group)&-&-&-& $0.63$ & $0.36$ & $293.54$  \\
     IE+RE+ID+AP(Wass)(group)&-&-&-& \tableSecond{0.75} & \tableSecond{0.48} & \tableSecond{291.16}\\
     \bottomrule
    \end{tabular}
    \label{tab:performance}
\end{table*}

{\bf Retrieval Methods Comparison.} Comparing {Retrieval}, {ATTEN\cite{chen2018deep}} and {ACME}\cite{wang2019learning}, our retrieval model achieves the best performance. Retrieval-based methods and prediction-based methods involving RE both use Food Space, but in different ways. Retrieval uses Food Space to find the nearest recipe neighbor, while prediction-based methods involving RE take Food Space features for further detection and estimation. Retrieval methods are all outperformed by {IE+RE+ID+AP}, which shows retrieval-based methods have their own limitations. For retrieval performance versus the number of recipes in the collection, please see our supplemental material.

{\bf Importance of RE (Food Space feature).} \autoref{tab:performance} demonstrates RE (Food Space feature) is the most important component for both ingredient detection and amount prediction. This is because IE learns to extract image features highly relevant to the corresponding recipe, including title and instructions information when training the retrieval model. The Food Space features are then used to condition ID and AP and the performances are improved.

{\bf Importance of ID.} 
\autoref{tab:performance} demonstrates ID improves {CVG} and {IOU} and only slightly degrades {EMD}. ID also provides information complementary to RE as {IE+RE+ID+AP} has the highest {CVG} and {IOU} and top-3 {EMD}. 

{\bf Role of Substitution Groups during Training.} {IE+RE+ID+AP(CE)} and {IE+RE+ID+AP(Wass)} have the same ID so they have the same {CVG} and {IOU}. {IE+RE+ID+AP(Wass)} utilizes substitution information when training AP but {IE+RE+ID+AP(CE)} does not. {IE+RE+ID+AP(Wass)} has a slightly better {EMD} because the model learns to predict results with larger optimal flows for ingredient pairs with smaller distances. However, compare {IE+RE+AP(Wass)} with {IE+RE+AP(CE)}, {IE+RE+AP(Wass)} utilizes substitution information during training while {IE+RE+AP(CE)} does not. The ingredient detection performance of {IE+RE+AP(Wass)} is severely degraded. This is because {IE+RE+AP(Wass)} tends to predict multiple ingredients from the same substitution group to match the amounts, see \autoref{sec:example}, which is not desired by users. Therefore, substitution information can improve AP but using it for ID still needs further research.

{\bf Ingredient-level vs Group-level Methods.} Ingredient-level methods predict detailed ingredient-level amounts while group-level methods view each group as a whole and can only predict group-level results. %What specific ingredients the results come from cannot be inferred.
\autoref{tab:performance} shows that {IOU (Group)} is always higher than the ingredient level {IOU} as predicting multiple ingredients within the same group is penalized for ingredient-level metrics but not for group-level metrics. 
%The group is viewed as predicted as long as at least one ingredient in the group is predicted. 
%as the number of common ingredients is the smaller number between the two for this group, result in a penalty when the two numbers are not equal. 
{EMD (Group)} is larger than ingredient level {EMD} because ingredient-level methods are trained to look for optimal transportations between close ingredient neighbors if the pair does not belong to the same group, which are much closer than cluster centroids used in {EMD (Group)}. Therefore, the distances become larger. {IE+RE+ID+AP(Wass)(group)} performs the best if only group-level metrics are needed.

\subsection{Results on Sample Images}
\label{sec:example}
\begin{figure*}[t!]
    \centering
    \begin{minipage}[t]{.2\linewidth}
        \centering
        \includegraphics[align=t,width=0.75\textwidth]{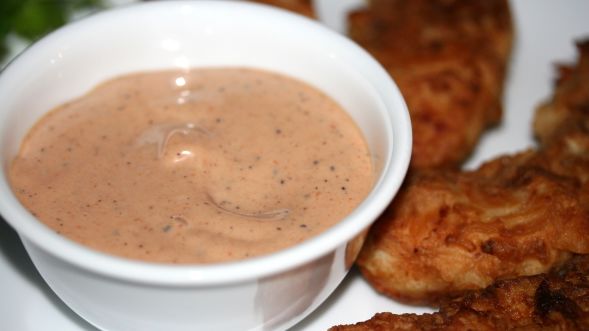}
        \caption*{\footnotesize \centering Cane Sauce (For Dippin' Chicken)}
    \end{minipage}
    \begin{minipage}[t]{.2\linewidth}
        \includegraphics[align=t,width=0.75\textwidth]{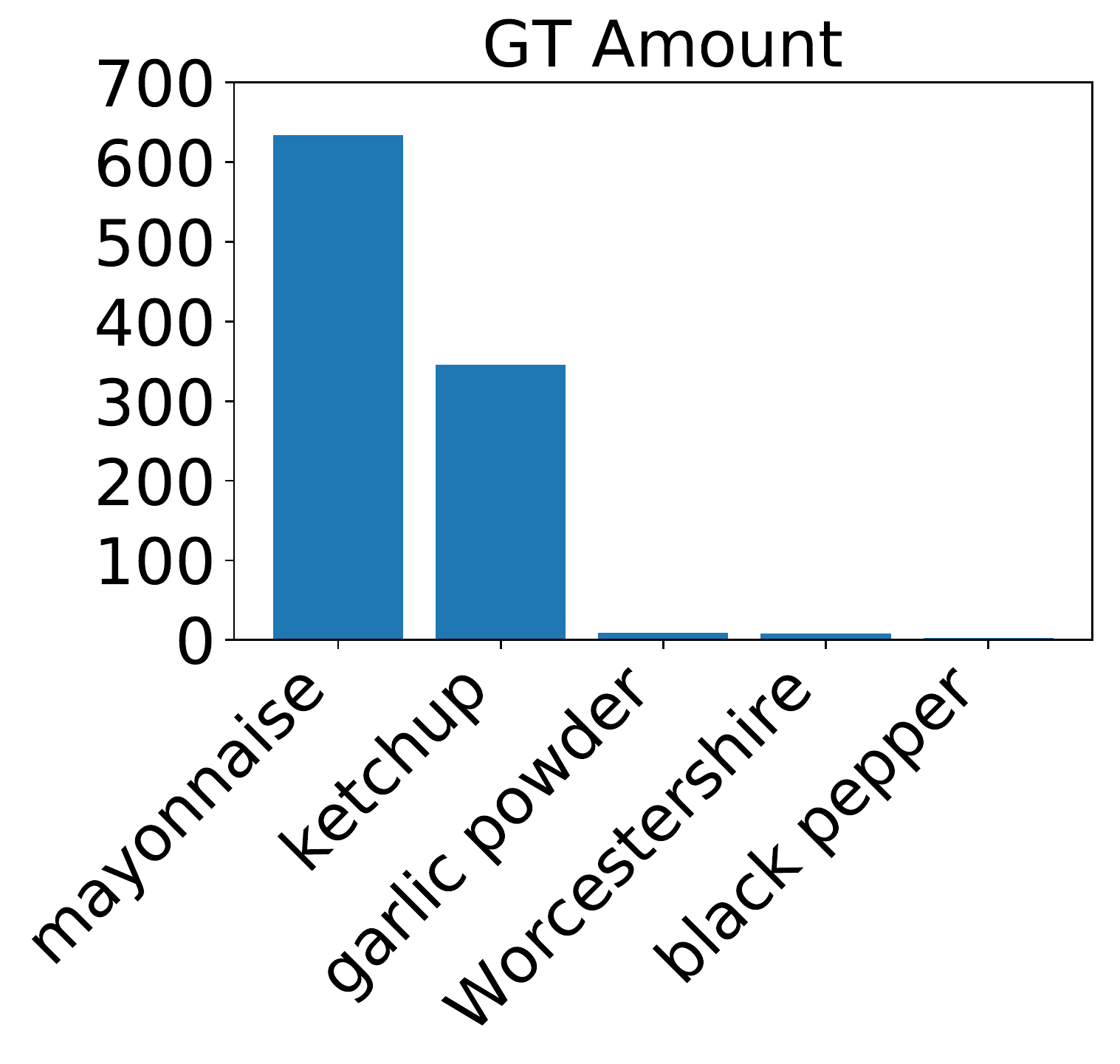}
    \end{minipage}
    \begin{minipage}[t]{.2\linewidth}
        \includegraphics[align=t,width=0.8\textwidth]{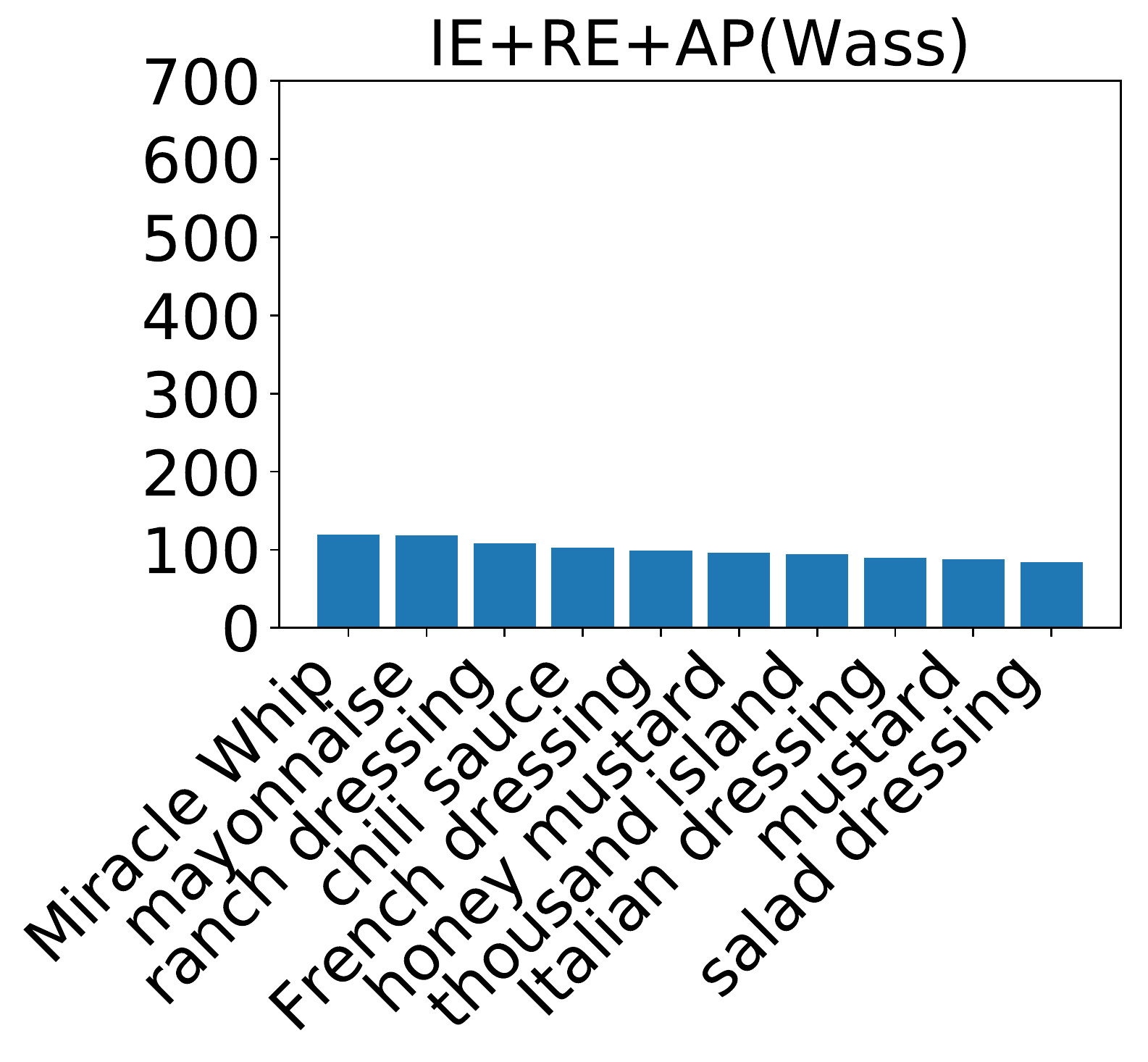}
    \end{minipage}
    \begin{minipage}[t]{.2\linewidth}
        \includegraphics[align=t,width=0.75\textwidth]{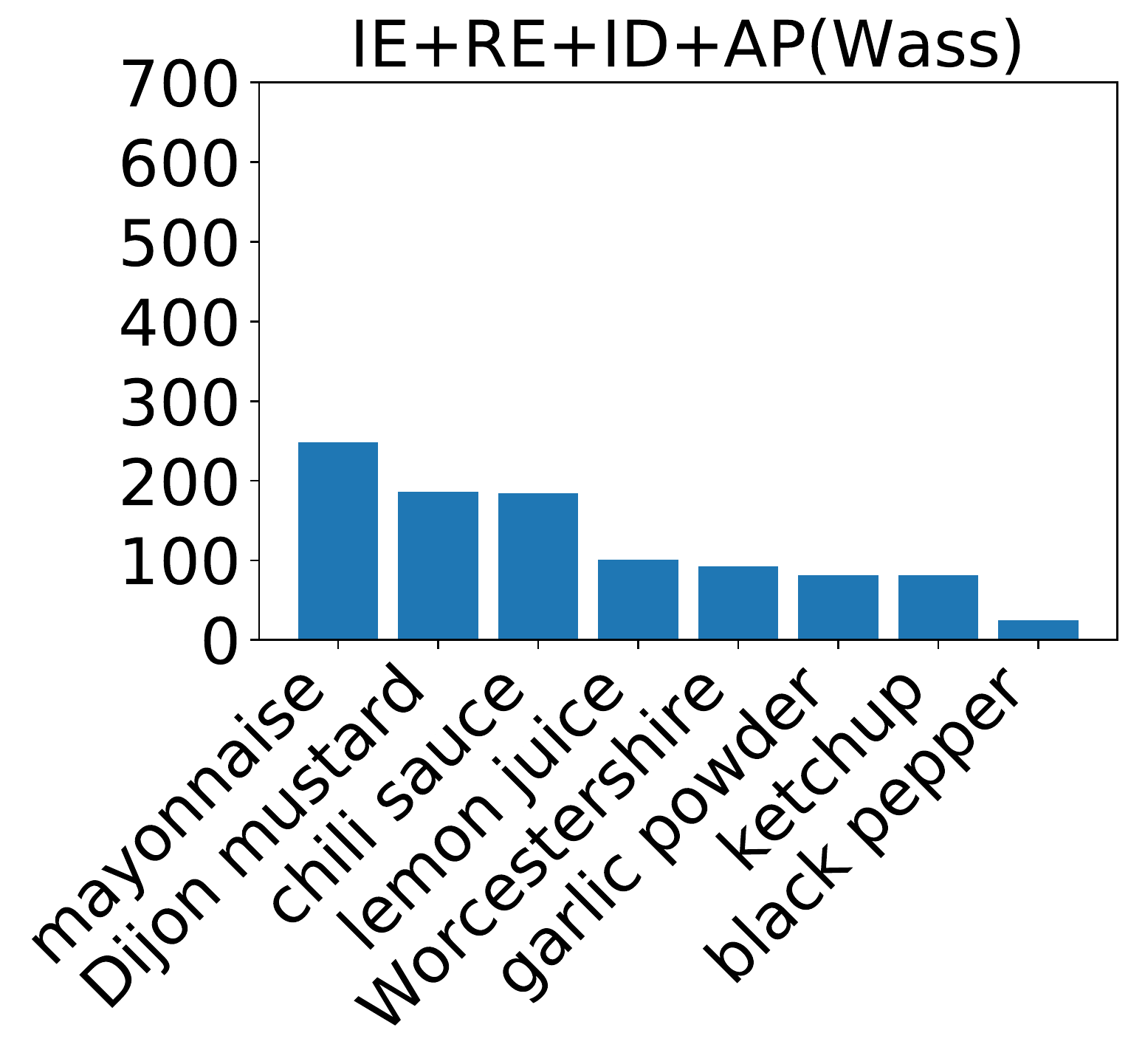}
    \end{minipage}
    \begin{minipage}[t]{.2\linewidth}
        \centering
        \includegraphics[align=t,width=0.75\textwidth]{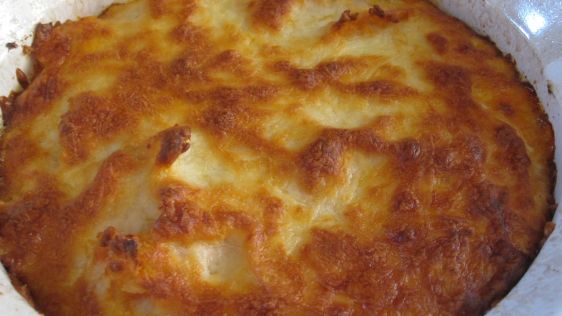}
        \caption*{\footnotesize \centering Creamy Baked Ziti}
    \end{minipage}
    \begin{minipage}[t]{.2\linewidth}
        \includegraphics[align=t,width=0.8\textwidth]{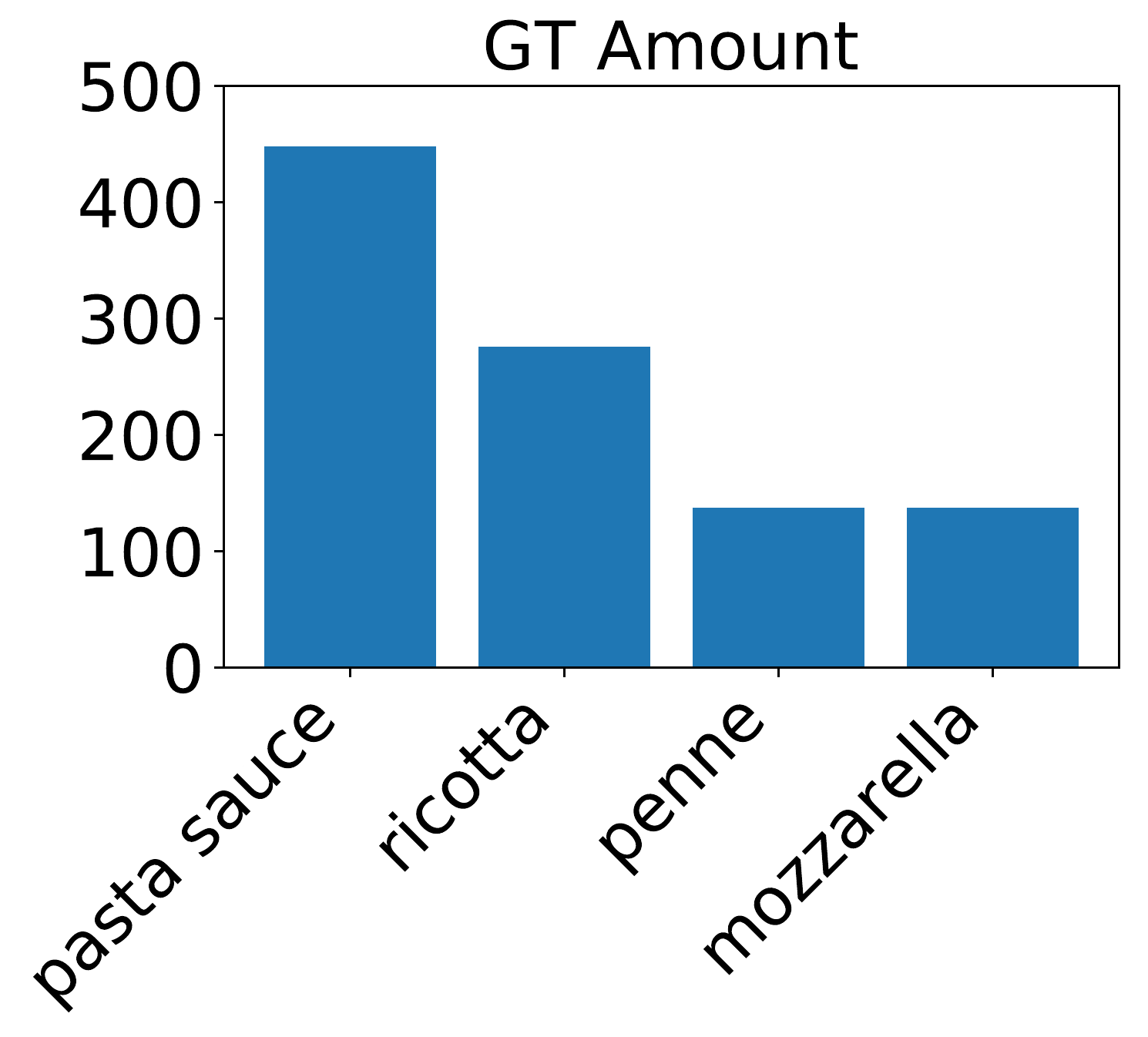}
    \end{minipage}
    \begin{minipage}[t]{.2\linewidth}
        \includegraphics[align=t,width=0.85\textwidth]{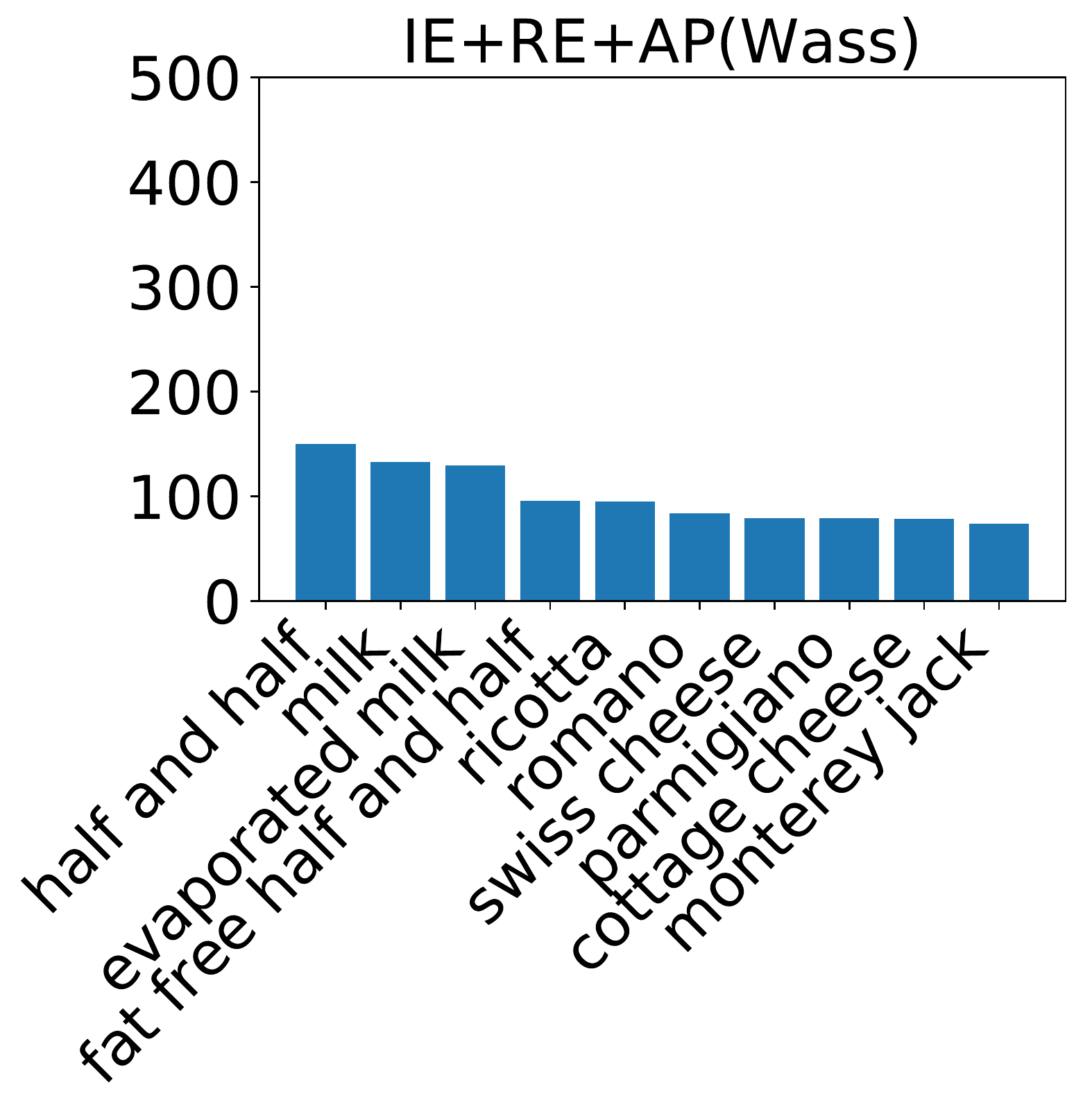}
    \end{minipage}
    \begin{minipage}[t]{.2\linewidth}
        \includegraphics[align=t,width=0.75\textwidth]{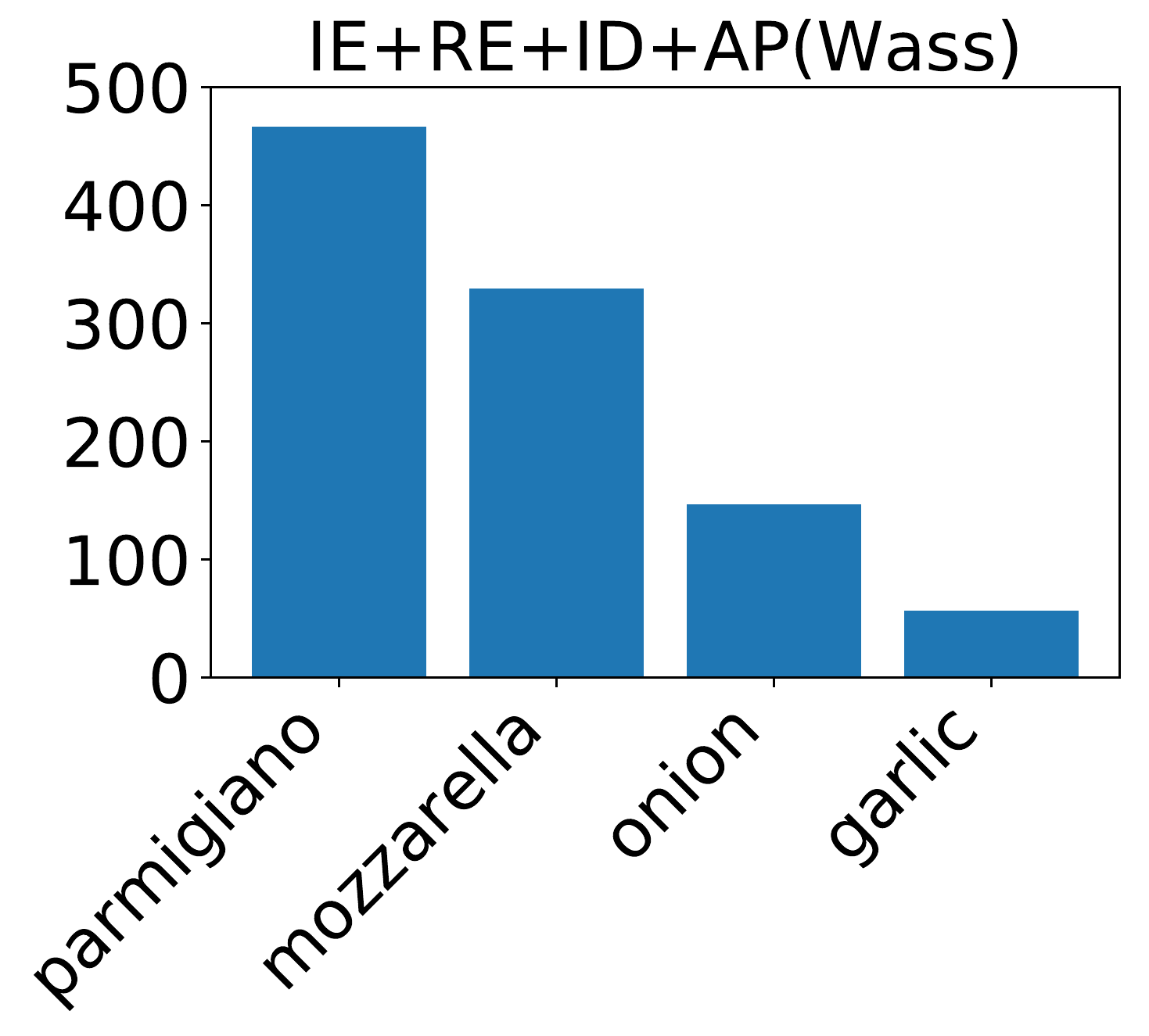}
    \end{minipage}
    \begin{minipage}[t]{.2\linewidth}
        \centering
        \includegraphics[align=t,width=0.75\textwidth]{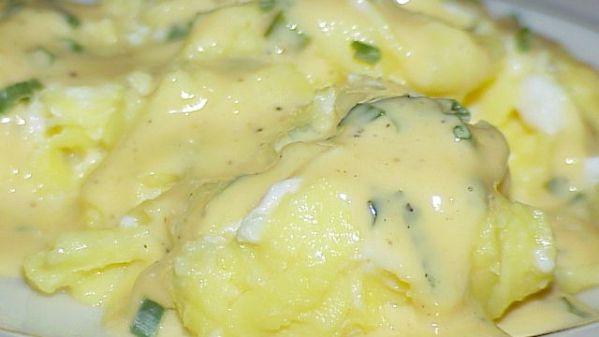}
        \caption*{\footnotesize \centering Scrambled Eggs}
    \end{minipage}
    \begin{minipage}[t]{.2\linewidth}
        \includegraphics[align=t,width=0.75\textwidth]{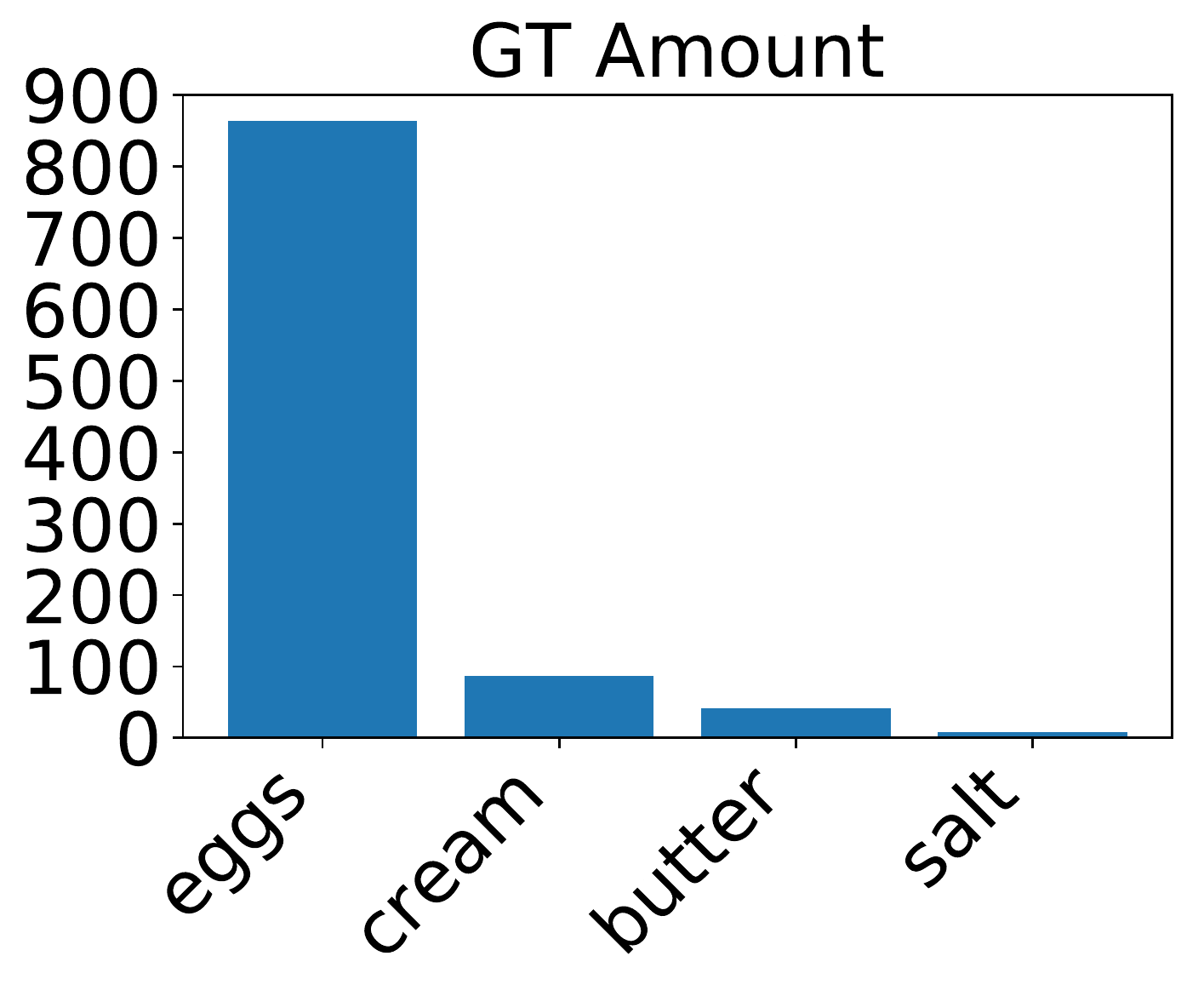}
    \end{minipage}
    \begin{minipage}[t]{.2\linewidth}
        \includegraphics[align=t,width=0.85\textwidth]{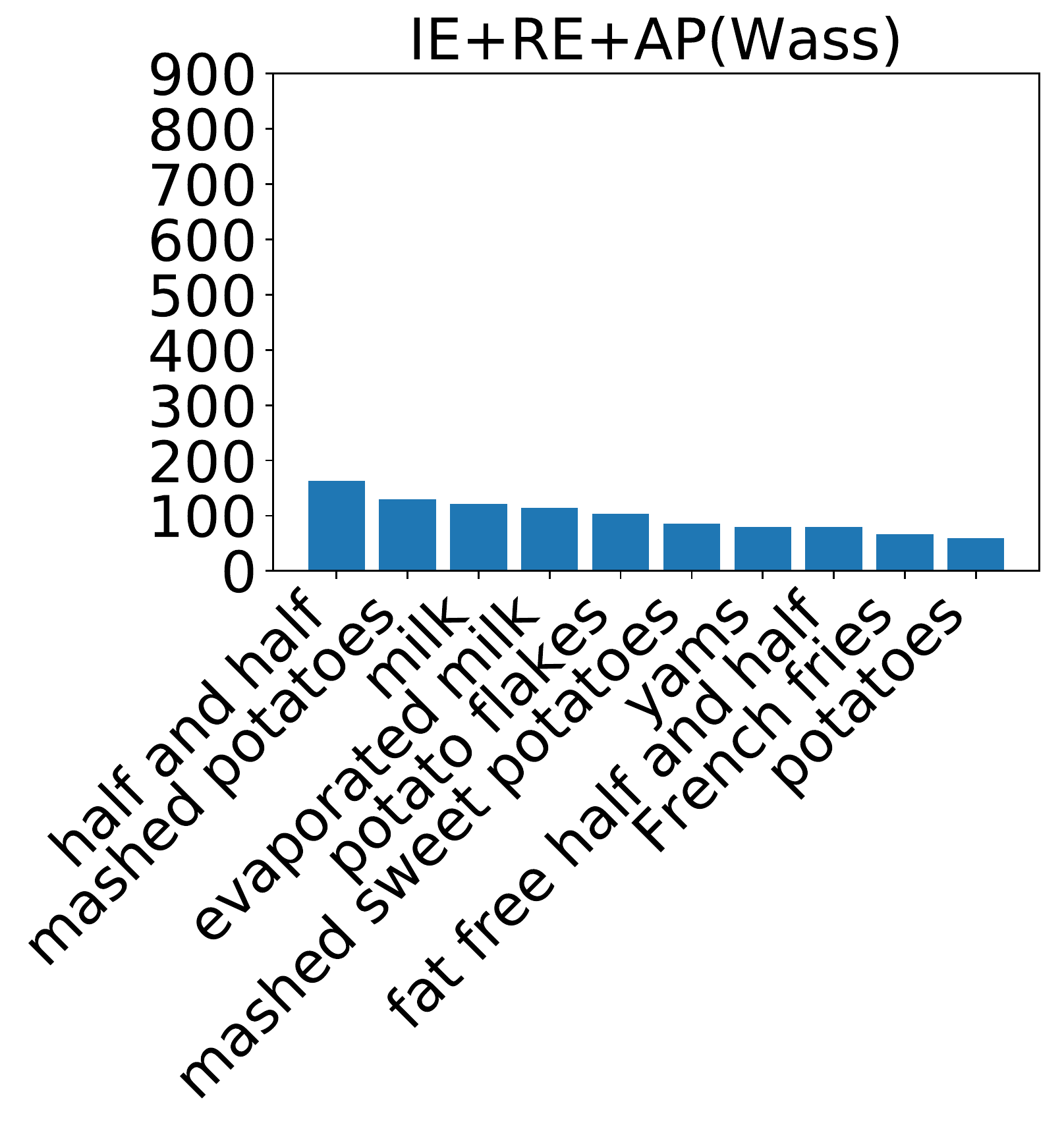}
    \end{minipage}
    \begin{minipage}[t]{.2\linewidth}
        \includegraphics[align=t,width=0.75\textwidth]{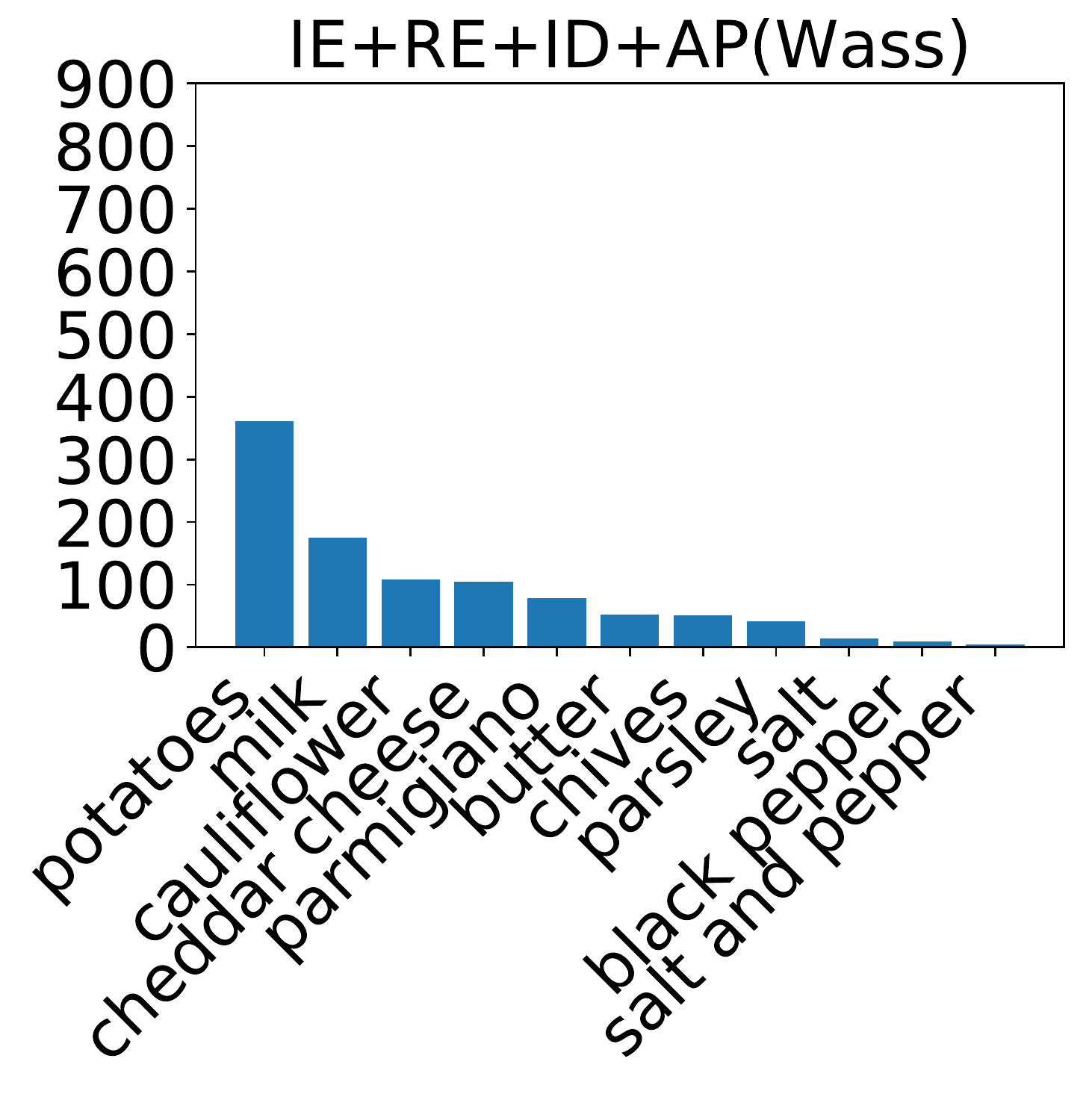}
    \end{minipage}
    %\caption{Easy, average and difficult examples. Top row: easy, middle row: average, bottom row: difficult. The amounts are normalized to $C=1000$.}
    \caption{The predictions of IE+RE+AP(Wass), IE+RE+ID+AP(Wass) and the ground truth amounts. Top row: IE+RE+ID+AP(Wass) performs much better than the average. Middle row: IE+RE+ID+AP(Wass) around the average. Bottom row: IE+RE+ID+AP(Wass) much worse than the average. The amounts are linearly scaled such that the total weight is $1000$ grams.  
    %\hl{Is there a particular reason to keep underscores in the ingredient names?  I would remove them.} 
    %\hl{If I think about these images, they would be super hard for someone (human) to estimate ingredient amounts.  It is almost surprising, then, that the top one is the case where the amounts are estimated most accurately.  What that tells me, perhaps, is that the top one is easy to recall (the recipe of) and that there are not too many similar recipes and similar looking images.  On the other hand, for the bottom one (eggs), there are so many dishes that look similar, so in the foodspace there are many similar dishes around that embedding.  Consequently, it is hard to pick the right recipe and its corresponding amounts to make accurate prediction.  Can we check the retrieval scores for these images?}
    }
    \label{fig:examples}
\end{figure*}

% We demonstrate the results of our proposed models on three examples, according to the performance of IE+RE+ID+AP(Wass). The model performs better than average on ``Cane Sauce (For Dippin’Chicken)'', about average for ``Creamy Baked Ziti'' and worse than average on ``Scrambled Eggs''. 
% % 
\autoref{fig:examples} highlights a cross-section of performance of the IE+RE+ID+AP(Wass) model. There, the top, middle and bottom rows, show cases of good, average and poor performance, respectively.
% 
% We compare the results of { IE+RE+AP(Wass)} and { IE+RE+ID+AP(Wass)} with ground truth amounts in \autoref{fig:examples}. 
Additionally, \autoref{fig:examples} displays a comparison of prediction results from {IE+RE+AP(Wass)} and {IE+RE+ID+AP(Wass)} with ground truth amounts. 
%In these three examples, {\bf IE+RE+ID+AP(Wass)}  around the average and much worse than the average in terms of {\bf IOU} and {\bf EMD}. 
%
% The evaluation metrics are in \autoref{tab:examples}. The better method is marked ${\mathbf {bold}}$. 
\autoref{tab:examples} gives the specific performance values for both these methods on the examples shown on \autoref{fig:examples}.
For optimal earth mover's distance flow visualization please see supplemental material. 
\begin{table}[]
    \centering
    \caption{Metrics of Examples in \autoref{fig:examples}.}
    \begin{tabular}{p{2.2cm}p{2.5cm}ccc}
    \toprule
    Example& Method & {\bf CVG} & {\bf IOU}& {\bf EMD}  \\
    \midrule
    Cane Sauce (For  & IE+RE+AP(Wass) & $0.20$ & $0.07$ & \tableSecond{40.06} \\
    Dippin' Chicken)& IE+RE+ID+AP(Wass) & \tableSecond{1.00} & \tableSecond{0.63} & $63.77$ \\
    \midrule
    Creamy Baked Ziti & IE+RE+AP(Wass) & \tableSecond{0.50} & $0.17$ & $192.33$ \\
    & IE+RE+ID+AP(Wass) & \tableSecond{0.50} & \tableSecond{0.33} & \tableSecond{158.62} \\
    \midrule
    Scrambled Eggs & IE+RE+AP(Wass) & $0.00$ & $0.00$ & \tableSecond{252.91} \\
    &  IE+RE+ID+AP(Wass) & \tableSecond{0.50} & \tableSecond{0.15} & $347.04$ \\
    \bottomrule
    \end{tabular}
    \label{tab:examples}
\end{table}

In Cane Sauce (For Dippin' Chicken), there is only one match in {IE+RE+AP(Wass)}, mayonnaise, on the other hand, every ingredient in ground truth is in {IE+RE+ID+AP(Wass)}. However, as the major ingredient in ground truth is mayonnaise, {IE+RE+AP(Wass)} predicts multiple similar salad dressings, trying to match the total amount of mayonnaise, resulting in a smaller {EMD}.  In Creamy Baked Ziti, the two cheeses in the ground truth are detected. {IE+RE+AP(Wass)} predicts more items in the cheese group, leading to a drop in {IOU}. As for amounts, the error in {IE+RE+AP(Wass)} mostly comes from milk and half-and-half, while the error in {IE+RE+ID+AP(Wass)} mostly comes from onion. In the Scrambled Eggs, none of the ingredients in ground truth is detected in {IE+RE+AP(Wass)} while butter and salt are detected in {IE+RE+ID+AP(Wass)}. But for amounts, butter and salt take up less than 100g in ground truth. For both methods, the amount error comes from matching other ingredients to eggs. {IE+RE+ID+AP(Wass)} has a larger error because of 100g cauliflower, which has a larger difference to eggs than other ingredients predicted by the two methods. However, the scrambled eggs in the image look creamy, therefore, it might be reasonable to predict it as mashed potatoes, possibly with cauliflower. 
% Even in the existence of examples difficult to predict, the methods still generate reasonable outputs.
This highlights the methods' ability to produce plausible results even when presented with challenging images.

\section{Conclusion}
% In this paper, we study a novel and challenging problem: g
We propose \ItoA, a deep learning architecture to predict the relative amount of each ingredient in a given food image.
% needed to prepare the observed food item. 
% We propose {\bf Image2Amounts} to solve the problem. More specifically, we 
We first learn an image embedding representation from a cross-modal retrieval systems. Next, we detect ingredients from ingredient detection networks and predict amounts conditioned on ingredient detection results using an amount prediction network. 
% We constructed 
Ingredient substitution groups are constructed to facilitate functional ingredient substitutions as a part of the evaluation metric and loss function. Experiments on annotated Recipe1M~\cite{salvador2017learning} show that our method generates state-of-the-art results, improving previous baselines. Even in the presence of challenging test examples, the model is still able to yield robust ingredient amount estimates.
\bibliographystyle{IEEEtran}
\bibliography{IEEEabrv}

\begin{thebibliography}{10}
\providecommand{\url}[1]{#1}
\csname url@samestyle\endcsname
\providecommand{\newblock}{\relax}
\providecommand{\bibinfo}[2]{#2}
\providecommand{\BIBentrySTDinterwordspacing}{\spaceskip=0pt\relax}
\providecommand{\BIBentryALTinterwordstretchfactor}{4}
\providecommand{\BIBentryALTinterwordspacing}{\spaceskip=\fontdimen2\font plus
\BIBentryALTinterwordstretchfactor\fontdimen3\font minus
  \fontdimen4\font\relax}
\providecommand{\BIBforeignlanguage}[2]{{%
\expandafter\ifx\csname l@#1\endcsname\relax
\typeout{** WARNING: IEEEtran.bst: No hyphenation pattern has been}%
\typeout{** loaded for the language `#1'. Using the pattern for}%
\typeout{** the default language instead.}%
\else
\language=\csname l@#1\endcsname
\fi
#2}}
\providecommand{\BIBdecl}{\relax}
\BIBdecl

\bibitem{beijbom2015menu}
O.~Beijbom, N.~Joshi, D.~Morris, S.~Saponas, and S.~Khullar, ``Menu-match:
  Restaurant-specific food logging from images,'' in \emph{IEEE Winter
  Conference on Applications of Computer Vision}.\hskip 1em plus 0.5em minus
  0.4em\relax IEEE, 2015, pp. 844--851.

\bibitem{meyers2015im2calories}
A.~Meyers, N.~Johnston, V.~Rathod, A.~Korattikara, A.~Gorban, N.~Silberman,
  S.~Guadarrama, G.~Papandreou, J.~Huang, and K.~P. Murphy, ``Im2calories:
  towards an automated mobile vision food diary,'' in \emph{Proceedings of the
  IEEE International Conference on Computer Vision}, 2015, pp. 1233--1241.

\bibitem{salvador2017learning}
A.~Salvador, N.~Hynes, Y.~Aytar, J.~Marin, F.~Ofli, I.~Weber, and A.~Torralba,
  ``Learning cross-modal embeddings for cooking recipes and food images,'' in
  \emph{Proceedings of the IEEE conference on computer vision and pattern
  recognition}, 2017, pp. 3020--3028.

\bibitem{chen2016deep}
J.~Chen and C.-W. Ngo, ``Deep-based ingredient recognition for cooking recipe
  retrieval,'' in \emph{Proceedings of the 24th ACM international conference on
  Multimedia}.\hskip 1em plus 0.5em minus 0.4em\relax ACM, 2016, pp. 32--41.

\bibitem{he2009comprehensive}
F.~J. He and G.~A. MacGregor, ``A comprehensive review on salt and health and
  current experience of worldwide salt reduction programmes,'' \emph{Journal of
  human hypertension}, vol.~23, no.~6, p. 363, 2009.

\bibitem{han2019art}
F.~Han, R.~Guerrero, and V.~Pavlovic, ``The art of food: Meal image synthesis
  from ingredients,'' \emph{arXiv preprint arXiv:1905.13149}, 2019.

\bibitem{mikolov2013distributed}
T.~Mikolov, I.~Sutskever, K.~Chen, G.~S. Corrado, and J.~Dean, ``Distributed
  representations of words and phrases and their compositionality,'' in
  \emph{Advances in neural information processing systems}, 2013, pp.
  3111--3119.

\bibitem{bossard2014food}
L.~Bossard, M.~Guillaumin, and L.~Van~Gool, ``Food-101--mining discriminative
  components with random forests,'' in \emph{European Conference on Computer
  Vision}.\hskip 1em plus 0.5em minus 0.4em\relax Springer, 2014, pp. 446--461.

\bibitem{ciocca2017food}
G.~Ciocca, P.~Napoletano, and R.~Schettini, ``Food recognition: A new dataset,
  experiments, and results.'' \emph{IEEE J. Biomedical and Health Informatics},
  vol.~21, no.~3, pp. 588--598, 2017.

\bibitem{wang2015recipe}
X.~Wang, D.~Kumar, N.~Thome, M.~Cord, and F.~Precioso, ``Recipe recognition
  with large multimodal food dataset,'' in \emph{2015 IEEE International
  Conference on Multimedia and Expo (ICME)}.\hskip 1em plus 0.5em minus
  0.4em\relax IEEE, 2015, pp. 1--6.

\bibitem{kagaya2014food}
H.~Kagaya, K.~Aizawa, and M.~Ogawa, ``Food detection and recognition using
  convolutional neural network,'' in \emph{Proceedings of the 22nd ACM
  international conference on Multimedia}, 2014, pp. 1085--1088.

\bibitem{singla2016food}
A.~Singla, L.~Yuan, and T.~Ebrahimi, ``Food/non-food image classification and
  food categorization using pre-trained googlenet model,'' in \emph{Proceedings
  of the 2nd International Workshop on Multimedia Assisted Dietary
  Management}.\hskip 1em plus 0.5em minus 0.4em\relax ACM, 2016, pp. 3--11.

\bibitem{merler2016snap}
M.~Merler, H.~Wu, R.~Uceda-Sosa, Q.-B. Nguyen, and J.~R. Smith, ``Snap, eat,
  repeat: a food recognition engine for dietary logging,'' in \emph{Proceedings
  of the 2nd international workshop on multimedia assisted dietary management},
  2016, pp. 31--40.

\bibitem{chen2017chinesefoodnet}
X.~Chen, Y.~Zhu, H.~Zhou, L.~Diao, and D.~Wang, ``Chinesefoodnet: A large-scale
  image dataset for chinese food recognition,'' \emph{arXiv preprint
  arXiv:1705.02743}, 2017.

\bibitem{mezgec2017nutrinet}
S.~Mezgec and B.~Korou{\v{s}}i{\'c}~Seljak, ``Nutrinet: a deep learning food
  and drink image recognition system for dietary assessment,''
  \emph{Nutrients}, vol.~9, no.~7, p. 657, 2017.

\bibitem{martinel2018wide}
N.~Martinel, G.~L. Foresti, and C.~Micheloni, ``Wide-slice residual networks
  for food recognition,'' in \emph{2018 IEEE Winter Conference on Applications
  of Computer Vision (WACV)}.\hskip 1em plus 0.5em minus 0.4em\relax IEEE,
  2018, pp. 567--576.

\bibitem{qiu2019mining}
J.~Qiu, F.~P.~W. Lo, Y.~Sun, S.~Wang, and B.~Lo, ``Mining discriminative food
  regions for accurate food recognition.'' in \emph{BMVC}, 2019, p. 158.

\bibitem{chen2017cross}
J.~Chen, L.~Pang, and C.-W. Ngo, ``Cross-modal recipe retrieval: How to cook
  this dish?'' in \emph{International Conference on Multimedia Modeling}.\hskip
  1em plus 0.5em minus 0.4em\relax Springer, 2017, pp. 588--600.

\bibitem{carvalho2018cross}
M.~Carvalho, R.~Cad{\`e}ne, D.~Picard, L.~Soulier, N.~Thome, and M.~Cord,
  ``Cross-modal retrieval in the cooking context: Learning semantic text-image
  embeddings,'' in \emph{The 41st International ACM SIGIR Conference on
  Research \& Development in Information Retrieval}, 2018, pp. 35--44.

\bibitem{chen2018deep}
J.-J. Chen, C.-W. Ngo, F.-L. Feng, and T.-S. Chua, ``Deep understanding of
  cooking procedure for cross-modal recipe retrieval,'' in \emph{Proceedings of
  the 26th ACM international conference on Multimedia}.\hskip 1em plus 0.5em
  minus 0.4em\relax ACM, 2018, pp. 1020--1028.

\bibitem{wang2019learning}
H.~Wang, D.~Sahoo, C.~Liu, E.-p. Lim, and S.~C. Hoi, ``Learning cross-modal
  embeddings with adversarial networks for cooking recipes and food images,''
  in \emph{Proceedings of the IEEE Conference on Computer Vision and Pattern
  Recognition}, 2019, pp. 11\,572--11\,581.

\bibitem{salvador2019inverse}
A.~Salvador, M.~Drozdzal, X.~Giro-i Nieto, and A.~Romero, ``Inverse cooking:
  Recipe generation from food images,'' in \emph{Proceedings of the IEEE
  Conference on Computer Vision and Pattern Recognition}, 2019, pp.
  10\,453--10\,462.

\bibitem{min2019ingredient}
W.~Min, L.~Liu, Z.~Luo, and S.~Jiang, ``Ingredient-guided cascaded
  multi-attention network for food recognition,'' in \emph{Proceedings of the
  27th ACM International Conference on Multimedia}, 2019, pp. 1331--1339.

\bibitem{shashua1994relative}
A.~Shashua and N.~Navab, ``Relative affine structure: Theory and application to
  3d reconstruction from perspective views,'' in \emph{Proceedings of the IEEE
  Conference on Computer Vision and Pattern Recognition}, vol.~94, 1994, pp.
  483--489.

\bibitem{dehais2017two}
J.~Dehais, M.~Anthimopoulos, S.~Shevchik, and S.~Mougiakakou, ``Two-view 3d
  reconstruction for food volume estimation,'' \emph{IEEE transactions on
  multimedia}, vol.~19, no.~5, pp. 1090--1099, 2017.

\bibitem{zheng2018multi}
X.~Zheng, Y.~Gong, Q.~Lei, R.~Yao, and Q.~Yin, ``Multi-view model contour
  matching based food volume estimation,'' in \emph{International Conference on
  Applied Human Factors and Ergonomics}.\hskip 1em plus 0.5em minus 0.4em\relax
  Springer, 2018, pp. 85--93.

\bibitem{liang2017computer}
Y.~Liang and J.~Li, ``Computer vision-based food calorie estimation: dataset,
  method, and experiment,'' \emph{arXiv preprint arXiv:1705.07632}, 2017.

\bibitem{fang2018single}
S.~Fang, Z.~Shao, R.~Mao, C.~Fu, E.~J. Delp, F.~Zhu, D.~A. Kerr, and C.~J.
  Boushey, ``Single-view food portion estimation: learning image-to-energy
  mappings using generative adversarial networks,'' in \emph{2018 25th IEEE
  International Conference on Image Processing (ICIP)}.\hskip 1em plus 0.5em
  minus 0.4em\relax IEEE, 2018, pp. 251--255.

\bibitem{ege2019image}
T.~Ege, Y.~Ando, R.~Tanno, W.~Shimoda, and K.~Yanai, ``Image-based estimation
  of real food size for accurate food calorie estimation,'' in \emph{2019 IEEE
  Conference on Multimedia Information Processing and Retrieval (MIPR)}.\hskip
  1em plus 0.5em minus 0.4em\relax IEEE, 2019, pp. 274--279.

\bibitem{li2019deep}
J.~Li, R.~Guerrero, and V.~Pavlovic, ``Deep cooking: Predicting relative food
  ingredient amounts from images,'' in \emph{Proceedings of the 5th
  International Workshop on Multimedia Assisted Dietary Management}.\hskip 1em
  plus 0.5em minus 0.4em\relax ACM, 2019, pp. 2--6.

\bibitem{he2016deep}
K.~He, X.~Zhang, S.~Ren, and J.~Sun, ``Deep residual learning for image
  recognition,'' in \emph{Proceedings of the IEEE conference on computer vision
  and pattern recognition}, 2016, pp. 770--778.

\bibitem{hermans2017defense}
A.~Hermans, L.~Beyer, and B.~Leibe, ``In defense of the triplet loss for person
  re-identification,'' \emph{arXiv preprint arXiv:1703.07737}, 2017.

\bibitem{frogner2015learning}
C.~Frogner, C.~Zhang, H.~Mobahi, M.~Araya, and T.~A. Poggio, ``Learning with a
  wasserstein loss,'' in \emph{Advances in Neural Information Processing
  Systems}, 2015, pp. 2053--2061.

\end{thebibliography}

\end{document}